\documentclass[lettersize,journal]{IEEEtran}
\usepackage{amsmath,amsfonts}
\usepackage{array}
\usepackage{textcomp}
\usepackage{stfloats}
\usepackage{url}
\usepackage{verbatim}
\usepackage{graphicx}
\usepackage{subfigure}
\usepackage{cite}
\usepackage{booktabs}
\usepackage{adjustbox}

\usepackage{algorithm}
\usepackage{algorithmicx}
\usepackage{algpseudocode}
\usepackage{xcolor}
\usepackage{cite}
\usepackage{amsmath}
\usepackage{siunitx}

\usepackage{array}
\usepackage{colortbl}
\usepackage{arydshln}
\usepackage{multirow}
\usepackage{multicol}
\usepackage{dashrule}
\usepackage{fancyhdr}
\usepackage{bm}
\usepackage{color}
\usepackage{colortbl}
\definecolor{mygray}{gray}{.9}
\begin{document}

\title{Prototype-Guided Curriculum Learning for Zero-Shot Learning}

\author{Lei~Wang, Shiming~Chen, Guo-Sen~Xie, Ziming~Hong, Chaojian~Yu, Qinmu~Peng \\ and Xinge~You,~\IEEEmembership{Senior Member,~IEEE}

\thanks{Lei~Wang is with the School of Electronic Information and Communications, Huazhong University of Science and Technology, Wuhan, China.}

\thanks{Shiming~Chen is with the Mohamed bin Zayed University of AI, Abu Dhabi, United Arab Emirates.}%

\thanks{Guo-Sen~Xie is with the School of Computer Science and Engineering, Nanjing University of Science and Technology, Nanjing, China.}%

\thanks{Ziming Hong is with the School of Computer Science, The University of Sydney, Sydney, Australia.}

\thanks{Chaojian~Yu is with the School of Electronic Information and Communications, Huazhong University of Science and Technology, Wuhan, China.}

\thanks{Qinmu~Peng (corresponding author, e-mail: pengqinmu@hust.edu.cn) is with the School of Electronic Information and Communications, Huazhong University of Science and Technology, Wuhan, China.}

\thanks{Xinge~You is with the School of Electronic Information and Communications, Huazhong University of Science and Technology, Wuhan, China.}
}


\maketitle

\begin{abstract}
In Zero-Shot Learning (ZSL), embedding-based methods enable knowledge transfer from seen to unseen classes by learning a visual-semantic mapping from seen-class images to class-level semantic prototypes (e.g., attributes). However, these semantic prototypes are manually defined and may introduce noisy supervision for two main reasons: (i) instance-level mismatch: variations in perspective, occlusion, and annotation bias will cause discrepancies between individual sample and the class-level semantic prototypes; and (ii) class-level imprecision: the manually defined semantic prototypes may not accurately reflect the true semantics of the class. Consequently, the visual-semantic mapping will be misled, reducing the effectiveness of knowledge transfer to unseen classes. In this work, we propose a prototype-guided curriculum learning framework (dubbed as CLZSL), which mitigates instance-level mismatches through a Prototype-Guided Curriculum Learning (PCL) module and addresses class-level imprecision via a Prototype Update (PUP) module. Specifically, the PCL module prioritizes samples with high cosine similarity between their visual mappings and the class-level semantic prototypes, and progressively advances to less-aligned samples, thereby reducing the interference of instance-level mismatches to achieve accurate visual-semantic mapping. Besides, the PUP module dynamically updates the class-level semantic prototypes by leveraging the visual mappings learned from instances, thereby reducing class-level imprecision and further improving the visual-semantic mapping. Experiments were conducted on standard benchmark datasets—AWA2, SUN, and CUB—to verify the effectiveness of our method.
\end{abstract}


\section{Introduction}
\IEEEPARstart{Z}{ero}-shot learning (ZSL) enables the recognition of unseen classes by leveraging the intrinsic semantic relationships between seen and unseen classes \cite{Lampert2014AttributeBasedCF}. In ZSL, no training samples are available for the unseen classes that appear in the test set, and the label spaces of the training and test sets do not overlap. The knowledge transfer from seen to unseen classes is typically facilitated through the manually defined attributes (e.g., the attribute of birds ``Bill color orange" on CUB dataset \cite{Welinder2010CaltechUCSDB2}), which serve as intermediate information. The predefined attributes, identified as class-level semantic prototypes, are shared across samples of the same class. Depending on the scope of the label space during testing, ZSL can be categorized into conventional Zero-Shot Learning (CZSL) and generalized Zero-Shot Learning (GZSL). CZSL focuses solely on predicting unseen classes, whereas GZSL is capable of predicting both seen and unseen classes \cite{Huang2019GenerativeDA,Li2019LeveragingTI,pourpanah2022review}. ZSL has been widely applied in various fields, including image recognition, video content understanding, and cross-modal retrieval \cite{jayaraman2014zero,xu2025information,fu2015zero,zhang2016zero,li2025dcdl}.

\begin{figure}[t]
\small
\begin{center}
\includegraphics[width=8.5cm,height=4.8cm]{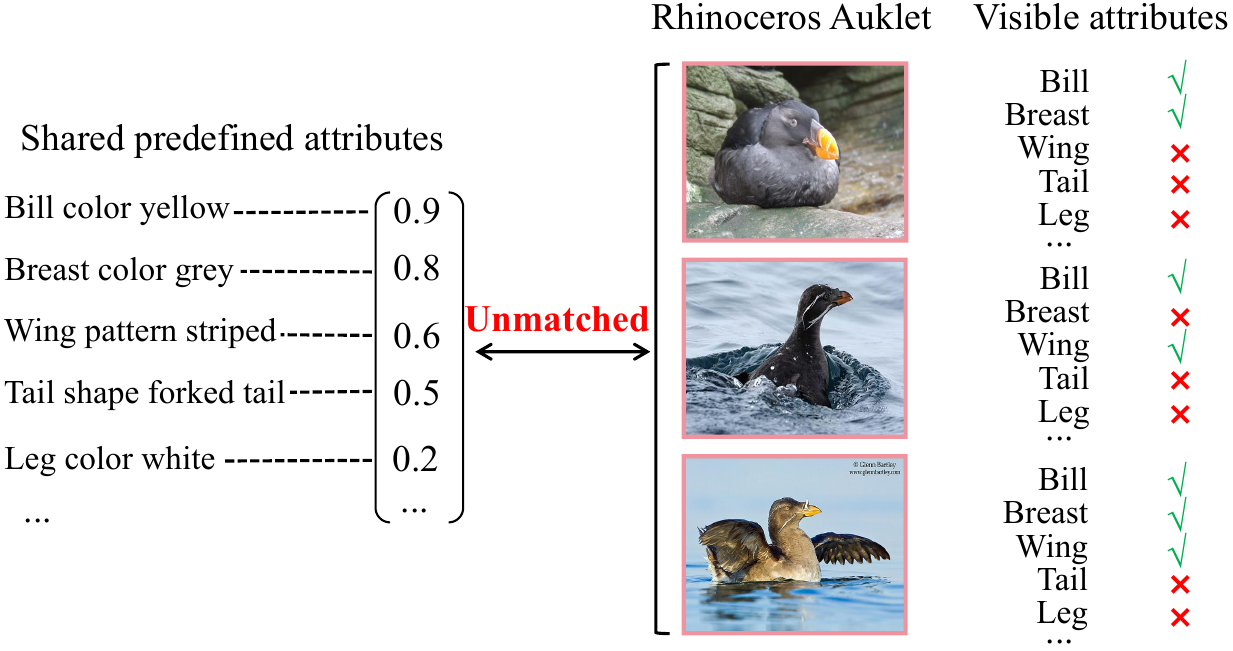}
\vspace{-2mm}
\caption{Motivation illustration. Due to the existence of perspective variations, occlusions and annotation biases, the class-level semantic prototypes (e.g., attributes) shared across samples of the same class do not accurately align with the instance-level semantic prototypes.} \vspace{-6mm}
\label{fig:fig1}
\end{center}
\end{figure}

In ZSL, effective knowledge transfer relies on learning an accurate mapping between visual and semantic \cite{xu2022attribute}. There are three main approaches to achieve the mapping between vision and semantics: embedding-based ZSL methods, generative ZSL methods and common subspace ZSL methods. Embedding-based ZSL methods learn a visual-semantic mapping by projecting visual features into the semantic space. \cite{akata2015label,Wang2017ZeroShotVR,Akata2015EvaluationOO}. Generative ZSL methods either map the attributes into the visual space to learn a semantic-visual mapping or convert ZSL into a traditional supervised learning task by generating synthetic samples of unseen classes using generative models such as GANs or VAEs \cite{Changpinyo2016SynthesizedCF,Yu2020EpisodeBasedPG,Chen2021FREE,hou2024visual,Xian2018FeatureGN,Xian2019FVAEGAND2AF}. Common subspace ZSL methods project both visual features and attributes into a shared embedding space to enable semantic alignment within that space \cite{Tsai2017LearningRV,Schnfeld2019GeneralizedZA}. Compared to the other two approaches, embedding-based ZSL methods offer explicit semantic mapping capabilities and can directly utilize rich semantic information, which often leads to better performance.

Early embedding-based ZSL methods utilize pre-trained models, such as ResNet-101 \cite{He2016DeepRL}, to extract global features from samples. These features are then mapped into a semantic space using implicit embedding methods in order to align with the attributes \cite{Song2018TransductiveUE,Li2018DiscriminativeLO}. Later approaches have improved upon this foundation by employing semantic attribute vectors—such as word embeddings extracted for each attribute using GloVe \cite{Pennington2014GloveGV}—to obtain region-specific features enriched with semantic information \cite{Xie2019AttentiveRE,Zhu2019SemanticGuidedML}. Additionally, some methods incorporate attention mechanisms to extract more refined visual features, enabling more precise alignment between the visual mappings and the attributes, which in turn improves the classification accuracy of ZSL models \cite{chen2022msdn,chen2022gndan,chen2022transzero}. 

Although these methods have improved ZSL performance, they often ignore the issue of \textit{noisy supervision}, which compromises the model's ability to learn precise visual-semantic mapping and thereby limits effective knowledge transfer between seen and unseen classes. This issue primarily arises from two aspects: 
\begin{itemize}
\item \textit{(i) Instance-level mismatch: variations in perspective, occlusion, and annotation bias can cause discrepancies between individual samples and their corresponding class-level semantic prototypes.} For example, as shown in Fig. \ref{fig:fig1}, the three images from the category ``Rhinoceros Auklet" in the CUB dataset are expected to share the same predefined attributes during training. However, certain attributes (e.g., ``tail") are invisible in these samples, resulting in discrepancies between these instances and their corresponding class-level semantic prototypes. \item \textit{(ii) Class-level imprecision: the manually defined semantic prototypes may not accurately reflect the true semantics of the class.} Class-level semantic prototypes are typically constructed through two distinct approaches. The first approach is based on cognitive science research, where human participants subjectively assess the strength of associations between attributes and classes based on semantic understanding, ultimately forming a class-level attribute matrix (e.g., AWA2 dataset \cite{Xian2017ZeroShotLC}). The second approach first annotating binary attributes for individual instances, and then computing the occurrence frequency of each attribute across instances within the same class to derive the final class-level attribute annotations (e.g., CUB dataset \cite{Welinder2010CaltechUCSDB2}). The former is prone to subjective bias, while the latter suffers from statistical errors, both leading to the manually defined semantic prototypes failing to accurately reflect the true semantics of the class.
\end{itemize}


To address these challenges in ZSL, we propose a prototype-guided curriculum learning framework (dubbed as CLZSL), which mitigates instance-level mismatches through a Prototype-Guided Curriculum Learning (PCL) module and addresses class-level imprecision via a Prototype Update (PUP) module. Specifically, the PCL module prioritizes samples with high cosine similarity between their visual mappings and the class-level semantic prototypes, and progressively advances to less-aligned samples. This prioritization is implemented by dynamically generating sample weights based on cosine similarity, where higher similarity scores yield higher weights, and vice versa. As training progresses, the model gradually uncovers more visual-semantic associations, leading to an increase in well-aligned samples and a decrease in less-aligned ones. This mechanism enables progressive learning, transitioning from well-aligned to less-aligned samples. As a result, the model reduces the interference of instance-level mismatches and improves the accuracy of visual-semantic mapping. Besides, the PUP module dynamically updates the class-level semantic prototypes by leveraging the visual mappings learned from instances, thereby reducing class-level imprecision and further enhancing the visual-semantic mapping. The two modules are alternately optimized to facilitate more effective knowledge transfer between seen and unseen classes, thereby improving the overall performance of the ZSL model.

To summarize, we make the following main contributions:
\begin{itemize}
\item We illustrate a key issue in ZSL: the class-level semantic prototypes (e.g., attributes) are manually defined and may introduce noisy supervision for two main reasons: (i) instance-level mismatch: variations in perspective, occlusion, and annotation bias will cause discrepancies between individual sample and the class-level semantic prototypes; and (ii) class-level imprecision: the manually defined semantic prototypes may not accurately reflect the true semantics of the class. Consequently, the visual-semantic mapping will be misled, reducing the effectiveness of knowledge transfer to unseen classes. \item We propose CLZSL, a novel framework with two core modules: PCL and PUP. PCL mitigates instance-level mismatches through a Prototype-Guided Curriculum Learning module and PUP addresses class-level imprecision via a Prototype Update module. By alternately optimizing the two modules, the model achieves more accurate visual-semantic mapping and effective knowledge transfer across seen and unseen classes, leading to improved ZSL performance. \item We conduct extensive comparative experiments and ablation studies on three widely used ZSL benchmark datasets AWA2, SUN, and CUB. The experimental results demonstrate the effectiveness of our approach in both CZSL and GZSL settings.
\end{itemize}

\section{Related Work}
\subsection{Zero-Shot Learning}
Early embedding-based ZSL methods primarily focus on learning the mapping between visual features and semantic attributes to enable knowledge transfer \cite{Song2018TransductiveUE,Li2018DiscriminativeLO,Yu2020EpisodeBasedPG,Min2020DomainAwareVB,Han2021ContrastiveEF,Chou2021AdaptiveAG,chen2025semantics}. These approaches typically extract global visual features using either pre-trained networks or end-to-end trainable architectures. However, they tend to overfit to seen classes, especially in GZSL setting. To address this issue, the generative ZSL methods utilize generative adversarial networks (GANs \cite{Goodfellow2014GenerativeAN}) or variational autoencoders (VAEs \cite{Mishra2018AGM}) to generate synthetic samples of unseen classes. This approach effectively transforms the ZSL problem into a traditional supervised recognition task, thereby reducing the domain bias between seen and unseen classes \cite{Sariyildiz2019GradientMG,Wu2020SelfSupervisedDG,Chen2020RethinkingGZ,Xie2021GeneralizedZL,chen2024causal,hu2025progressive}. End-to-end ZSL methods further improve performance by combining learning visual features and interactive embeddings jointly. Nevertheless, most of these methods rely solely on global visual features, often ignoring subtle inter-class differences \cite{Chen2022DUETCS,jiang2024dual}.

To better capture such fine-grained distinctions, attention-based ZSL methods \cite{Xie2019AttentiveRE,Zhu2019SemanticGuidedML,Xu2020AttributePN,wang2021dual} utilize semantic attribute descriptions (e.g., word embeddings of each semantic attribute extracted by GloVe) to identify more discriminative regional features. These approaches have significantly improved performance in the conventional ZSL (CZSL) setting. Recent works \cite{chen2022msdn,chen2022gndan,chen2022transzero,hong2022semantic,chen2022transzero++,cheng2023discriminative} have adopted more sophisticated architectures or enhanced attention mechanisms (e.g., transformer \cite{vaswani2017attention}) to achieve more effective visual-attribute interactions and learn more discriminative features. Despite these advancements, these methods still ignore a key issue: the class-level semantic prototypes (e.g., attributes) used in these models are manually defined, which may introduce noisy supervision. As a result, the visual-semantic mapping can be misled, reducing the effectiveness of knowledge transfer from seen to unseen classes and limiting further improvement of model performance.

\subsection{Curriculum Learning}
Curriculum learning is a training strategy that encourages the model to first learn from ``easy" samples before gradually progressing to more ``hard" ones \cite{jiang2015self}. This ``easy-to-hard" learning paradigm enables the model to capture underlying data patterns and avoid converging to suboptimal local solutions, thus enhancing model performance \cite{wang2021survey}. As the theory has evolved, the concept of curriculum learning has expanded beyond its original ``easy-to-hard" framework. It now includes a range of techniques for sample selection and weighting during training, from binary classification of sample difficulty to continuous importance weighting \cite{han2018co,jiang2018mentornet,soviany2022curriculum}.



Consider a classification problem with noisy labels. There are $n$ training samples $\{(\mathbf{x}_i,\mathbf{y}_i)\}^n_{i=1}$, where $\mathbf{x}_i$ denotes the $i^{th}$ input sample and $\mathbf{y}_i$ is its corresponding label. The overall optimization objective in curriculum learning can be formulated as follow:
\begin{gather}
\small
\label{eq:curriculum_learning}
\min_{\theta,\omega}H(\theta,\omega)
= \frac{1}{n}\sum_{i=1}^n \omega_i L(\mathbf{y}_i,g(\mathbf{x}_i,\theta)) + G(\omega_i; \varphi),
\end{gather}
\noindent where $L(\mathbf{y}_i,g(\mathbf{x}_i,\theta))$ is the loss function and $g(\mathbf{x}_i,\theta)$ is the discriminative function parameterized by $\theta$. The term $G(\omega_i; \varphi)$ represents the curriculum generation function, parameterized by $\varphi$, and $\omega_i$ is the weight assigned to the $i$-th training sample. In early approaches, the curriculum ranking was set manually, and knowledge transfer was achieved through incremental model training \cite{tsvetkov2016learning,el2020student}. Subsequently, automatic curriculum learning methods have been developed to dynamically adjust the difficulty order of training samples based on either the model's own feedback or external mechanisms \cite{fan2017self,ren2018self,platanios2019competence}.

Despite the diversity of existing curriculum learning approaches, there has been no prior exploration of applying curriculum learning theory to Zero-Shot recognition tasks. Our primary objective is to tackle the challenge that class-level semantic prototypes (e.g., attributes) used in these models are manually defined and may lead to instance-level mismatches.

\begin{figure*}[t]
\small
\begin{center}
\includegraphics[width=1\linewidth]{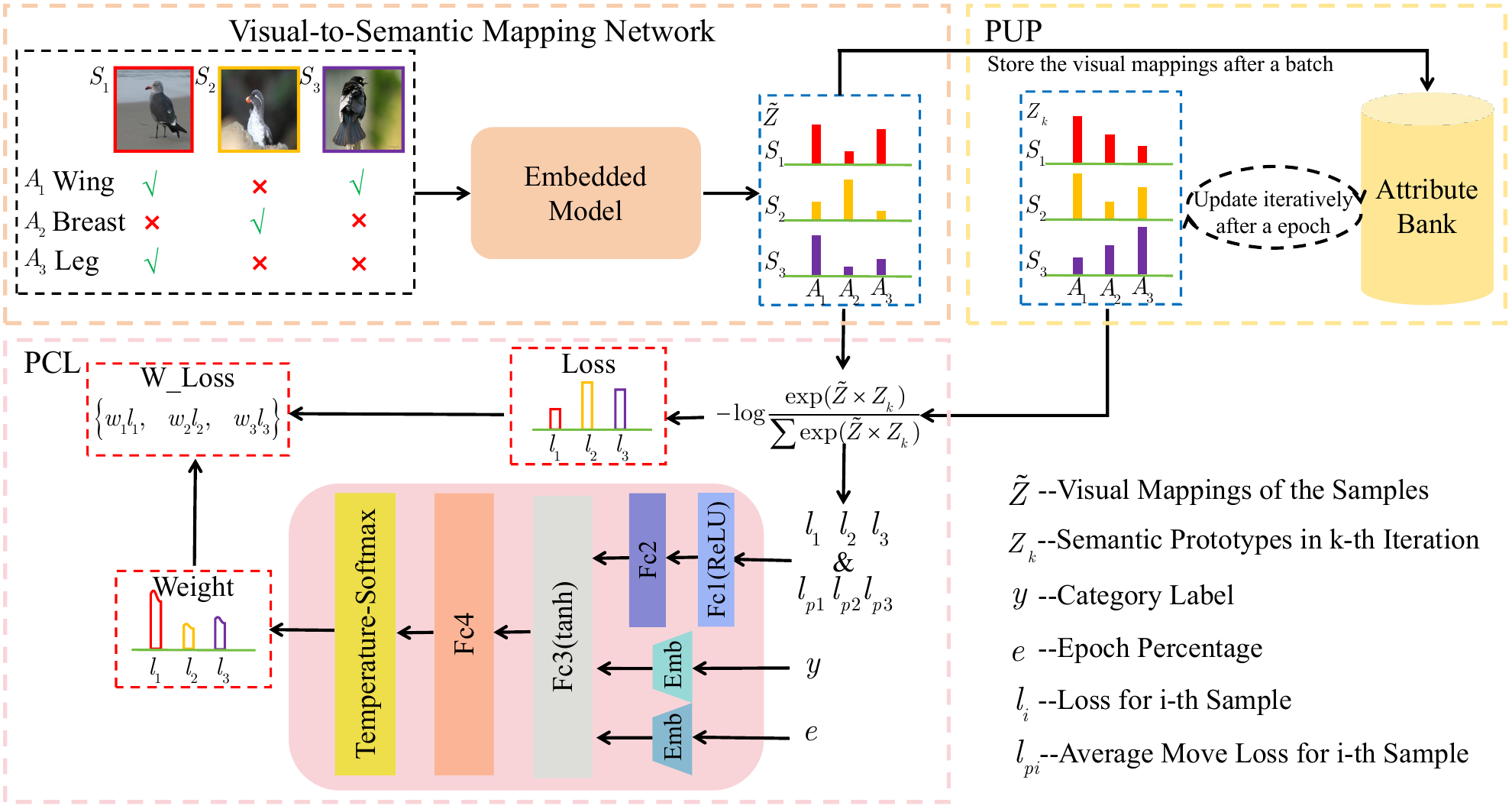}
\\\vspace{-2mm}
\caption{The overview of CLZSL. CLZSL comprises three parts: the visual-to-semantic mapping network, which serves as the basic component of embedding-based ZSL methods, and two core modules: the prototype-guided curriculum learning module (PCL) and the prototype update module (PUP).} \vspace{-6mm}
\label{fig:fig2}
\end{center}
\end{figure*}

\section{Proposed method}
We begin by introducing some notations and the problem definition. Assume that we have training data $\mathcal{D}^{s}=\left\{\left(x_{i}^{s}, y_{i}^{s}\right)\right\}^{N^s}_{i=1}$ with $C^s$ seen classes, where $x_i^s \in \mathcal{X}$ denotes the visual sample $i$, and $y_i^s \in \mathcal{Y}^s$ is the corresponding class label. Another set of unseen classes $C^u$ has unlabeled samples $\mathcal{D}^{u}=\left\{\left(x_{i}^{u}, y_{i}^{u}\right)\right\}^{N^u}_{i=1}$, where $x_{i}^{u}\in \mathcal{X}$ is the unseen class sample, and $y_{i}^{u} \in \mathcal{Y}^u$ is the corresponding label. The manually defined attributes (i.e., class-level semantic prototypes) $Z^{c}=\left[z_{1}^{c}, \ldots, z_{A}^{c}\right]^{\top}= \phi(y)$ for class $c \in \mathcal{C}^{s} \cup \mathcal{C}^{u} = \mathcal{C}$ with $|A|$ attributes help in knowledge transfer from seen to unseen classes. Note that we also use the semantic attribute vectors of each attribute $A=\{a_{1}, \ldots, a_{K}\}$ learned by GloVe \cite{Pennington2014GloveGV}. The aim of ZSL is to predict the class label $y_{i}^{u} \in \mathcal{Y}^u$ and $y_{i} \in \mathcal{Y} = \mathcal{Y}^s \cup \mathcal{Y}^u$ in CZSL and GZSL settings, respectively, where $\mathcal{Y}^s \cap \mathcal{Y}^u = \emptyset$.

In this paper, we propose a prototype-guided curriculum learning framework, dubbed as CLZSL, to address the issue that class-level semantic prototypes in ZSL are manually defined and may introduce noisy supervision, leading to ineffective knowledge transfer and hindering further improvement of model performance. As illustrated in Fig. \ref{fig:fig2}, the overview of CLZSL comprises three parts: the visual-to-semantic mapping network, which serves as the basic component of embedding-based ZSL methods, and two core modules: PCL and PUP. In the following section, we first introduce the basic component of embedding-based ZSL methods, followed by a detailed description of the two core modules, PCL and PUP.

\subsection{Visual-to-Semantic Mapping Network}
In ZSL, visual differences between categories are often concentrated in localized, discriminative regions. To address this, we identify and extract visual features from these local discriminative areas follow \cite{Huynh2020FineGrainedGZ}. In this way, semantic attribute vectors $A=\{a_{1}, \ldots, a_{K}\}$ (learned by GloVe) are leveraged to guide the extraction of attribute-specific visual features. This enables the model to learn more discriminative and semantically meaningful local features.

Specifically, given a set of visual features from an image, $V=\{v_1, \ldots, v_R\}$, where each $v_r$ encodes a distinct region of the image, the model performs attribute-specific attention by learning attention weights for the $a$-th attribute. These weights enable the model to attend to the most relevant regions of image $i$, as follows:
\begin{gather}
\small
\label{eq:attribute_weight}
\alpha_k^r = \frac{\exp \left(a_{k}^{\top} W_{1} v_{r}\right)}{\sum_{k=1}^{K} \exp \left(a_{k}^{\top} W_{1} v_{r}\right)},
\end{gather}
\noindent where $W_{1}$ is a learnable matrix used to compute the visual features of each region and measure the similarity between these features and the semantic attribute vectors. As a result, we obtain a set of attention weights $\{\alpha_k^r\}_{r=1}^{R}$ for each attribute. Using these attention weights, we then extract attribute-specific visual features for each semantic attribute as follow:
\begin{gather}
\small
\label{eq:regional_feature}
F_k = \sum_{r=1}^{R} \alpha_k^r v_r.
\end{gather}

After extracting the attribute-based visual features, we further introduce a mapping function $\mathcal{M}(\cdot)$ to project these features into the semantic embedding space. Specifically, $\mathcal{M}(\cdot)$ aligns the attribute-based visual features $F_k$ with their corresponding semantic attribute vectors $a_k$, formulated as:
\begin{gather}
\small
\label{eq:semantic_score}
\psi_k =\mathcal{M}(F_k)= a_{k}^{\top} W_{2} F_k,
\end{gather}
\noindent where $W_{2}$ is another learnable matrix. In fact, $\psi_k$ represents the attribute scores in an given image, indicating the confidence that image $i$ possesses the $k$-th attribute. We then obtain a mapped semantic embeddings $\psi(x)=\{\psi_1, \cdots,\psi_k\}$ for each input image. Lastly, we compute the class score $s^{c}$ as the product between the semantic embeddings $\psi(x_i)$ and the manually defined attributes $Z^{c}$, formulated as:
\begin{gather}
\small
\label{eq:class_score}
s^{c}=\psi(x_i) \times Z^{c}.
\end{gather}

Finally, we optimize the cross-entropy loss between the model's predictions and the ground-truth labels across all training images:
\begin{gather}
\small
\label{eq:l_ce_sc}\
\begin{aligned}
\mathcal{L}_{\text{CE}} =-\frac{1}{n} \sum_{i=1}^{n} [\log \frac{\exp \left(\psi(x_i^s) \times Z^{c}\right)}{\sum_{\hat{c} \in \mathcal{C}^s} \exp \left(\psi(x_i^s) \times Z^{\hat{c}} \right)}] .
\end{aligned}
\end{gather}

\subsection{Prototype-Guided Curriculum Learning Module}
Due to instance-level mismatches between individual samples and class-level semantic prototypes, the visual-semantic mapping can be misled, reducing the effectiveness of knowledge transfer to unseen classes. Therefore, we propose the PCL module. The PCL module dynamically generates the weights of training samples based on a carefully designed curriculum. In the early stages of training, the model prioritizes a small number of samples that are well-aligned with the class-level semantic prototypes and assigns them higher weights. As training progresses, the model gradually uncovers more visual-semantic associations, leading to an increase in well-aligned samples and a reduction in less-aligned samples. This process enables progressive learning from well-aligned to less-aligned samples. It is important to note that, unlike traditional curriculum learning methods \cite{tsvetkov2016learning,el2020student}, our approach is guided by semantic prototypes. Prototype-guided curriculum learning merges progressive learning with semantic prototypes. It models more precise interaction between visual features and semantic prototypes, improving mapping accuracy and enhancing knowledge transfer from seen to unseen classes.

Specifically, we obtain the similarity measurement between the semantic embeddings $\psi(x)$ and the manually defined attributes $Z^{c}$ for each sample from the visual-to-semantic mapping network. This result is represented as a loss vector $l_{bi}=\{l_{i}\}^b_{i=1}$, where b denotes the number of samples in a mini-batch. For stable learning and dynamic sample weight adjustment, we utilize the difference between the current sample loss $l_i$ and an exponentially moving average threshold $l_p$, as defined in Eq. \ref{eq:lp}, where $l_p$ is computed as the $p$-th percentile of the historical loss vector $\{l_{i}\}^b_{i=1}$. This method mitigates the impact of batch outliers and ensures consistent weight adjustments.
\begin{gather}
\small
\label{eq:lp}\
\begin{aligned}
l_{pi} = \{l_{i} - l_{p}\}^b_{i=1}.
\end{aligned}
\end{gather}

We then concatenate $l_{bi}$ and $l_{pi}$ and feed them into a two-layer fully connected neural network, with ReLU applied as the activation function to model nonlinear transformations:
\begin{gather}
\small
\label{eq:l_lp}\
\begin{aligned}
\delta_i= Fc_2(\sigma(Fc_1(Concat(l_{bi};l_{pi})))),
\end{aligned}
\end{gather}
\noindent where $Fc_1(\cdot)$ and $Fc_2(\cdot)$ denote the fully connected neural network, $\sigma(\cdot)$ represents the ReLU activation function, and $Concat(\cdot)$ denotes the concatenation operation.

We further incorporate additional ground truth labels and epoch information to enable more effective dynamic adjustment of sample weights. This is motivated by the observation that the model's understanding of the data evolves during training, necessitating a mechanism that adaptively responds to the training stage and prioritizes relevant samples. By incorporating epoch information, the model is able to dynamically adjust its focus on samples with varying levels of difficulty throughout the training process. Furthermore, incorporating label information enables the model to better assign higher weights to well-aligned samples while identifying less-aligned ones—such as those exhibiting low cosine similarity between their visual mappings and the class-level semantic prototypes. This facilitates a reduction in their influence and helps prevent overfitting to such mismatched samples.

Specifically, the class labels and the percentage of training completed (epoch percentage) are encoded using two separate embedding layers. As shown in the PCL module of Fig. \ref{fig:fig2}, the outputs of these embedding layers, together with the vector $\delta_i$ obtained from Eq. \ref{eq:l_lp}, are concatenated and fed into another pair of fully connected layers. The tanh activation function is applied between these layers to introduce nonlinear transformation:
\begin{gather}
\small
\label{eq:pcl}\
\begin{aligned}
\omega_i= \frac{exp(Fc_4(\rho(Fc_3(Concat(\epsilon_1(y_i);\epsilon_2(e);\delta_i))))/T)}{\sum^b_{i=1} exp(Fc_4(\rho(Fc_3(Concat(\epsilon_1(y_i);\epsilon_2(e);\delta_i))))/T)},
\end{aligned}
\end{gather}
\noindent where $\epsilon_1$ and $\epsilon_2$ denote the embedding layer for the class labels and the training epoch percentage, respectively. $Fc_3(\cdot)$ and $Fc_4(\cdot)$ represent the fully connected neural network. $\rho(\cdot)$ denotes the tanh activation function, and $Concat(\cdot)$ indicates the concatenation operation. To obtain the final sample weights, we apply temperature-softmax to map the output values to the range [0, 1]. The temperature parameter $T$ controls the smoothness of the weight distribution. A value $T < 1$ sharpens the distribution, emphasizing a few samples, while $T > 1$ produces smoother weights across samples. We set $T > 1$ in experiments to avoid over-reliance on low-loss samples.

\subsection{Prototype Update Module}
To address class-level imprecision, we propose the PUP module. The PUP module dynamically updates the class-level semantic prototypes by leveraging the visual mappings learned from instances. Specifically, during training, we introduce an additional branch that maps the visual features of samples into the semantic space. This mapping structure is identical to the visual-to-semantic mapping network. We maintain a dictionary that stores $\mathcal{C}$ key-value pairs ($\mathcal{C}$ being the total number of classes). Each key-value pair corresponds to a specific class and maintains the visual mappings of all images within that class, which are generated by the auxiliary mapping branch. For a given class with $\mathcal{N}$ samples, the corresponding value is a matrix of dimensions $\mathcal{N} \times \mathcal{A}$, where $\mathcal{A}$ denotes the total number of attributes. After each training epoch, we compute the average of the visual mappings learned from all samples within the same seen class. This averaged result is then used to update the class-level semantic prototypes of the corresponding seen class through weighted summation:
\begin{gather}
\small
\label{eq:prototype_update_seen}
\widehat{Z}^{c}_{s} = \beta Z^{c}_{s} + (1-\beta)\frac{1}{n}\sum_{i=1}^{N} \tilde{Z}^{c}_{i},
\end{gather}
\noindent where $Z^{c}_{s}$ denotes the class-level semantic prototypes for seen class $c$, $\tilde{Z}^{c}$ represents the visual mappings learned from instances belonging to class $c$, and $\beta$ is a trade-off parameter balancing the two terms.

For unseen classes, since no training samples are available, we update their class-level semantic prototypes by leveraging the visual mappings learned from instances of seen classes. Specifically, we use the class-level semantic prototypes for all classes $Z^{c}$ ($c \in \mathcal{C}^{s} \cup \mathcal{C}^{u} = \mathcal{C}$), which form a matrix of dimensions $\mathcal{C} \times \mathcal{A}$ (where $\mathcal{C}$ is the total number of classes and $\mathcal{A}$ is the number of attributes). We compute the transpose of this matrix (dimensions $\mathcal{A} \times \mathcal{C}$) and multiply it with the original matrix to obtain a semantic similarity matrix of dimensions $\mathcal{C} \times \mathcal{C}$. Each row in this matrix represents the semantic similarity between one class and all others. Next, we sort the elements in each row and select the top $k$ seen classes that are most semantically similar to the target unseen class. Subsequently, the class-level semantic prototypes for unseen classes are updated by leveraging the visual mappings learned from instances of their $k$ nearest seen class neighbors:
\begin{gather}
\small
\label{eq:prototype_update_unseen}
\widehat{Z}^{c'}_{u} = \beta Z^{c'}_{u} + (1-\beta)\frac{1}{k}\sum_{\bar{Z}^{c}_{j} \in N_{k}(\tilde{Z}^{c})} \bar{Z}^{c}_{j},
\end{gather}
\noindent where $Z^{c'}_{u}$ denotes the class-level semantic prototypes for unseen class $c'$, and $N_{k}(\tilde{Z}^{c})$ represents the set of $k$ nearest neighbors of ${Z}^{c'}_{u}$ in $\tilde{Z}^{c}$. $\beta$ is the trade-off parameter. We update the class-level semantic prototypes of both seen and unseen classes in equal proportion. At the beginning of training, the model has not yet learned an accurate mapping between visual and semantic, so we do not perform prototype update during the initial epochs. Instead, we start updating the prototypes only after the model has undergone a certain number of training epochs to ensure more reliable visual-semantic mapping. It is important to note that the updated class-level semantic prototypes are also used during the testing phase.

\begin{algorithm}[t]
\caption{The optimization of CLZSL.}
\label{algotithm:CLZSL_optimization}
\begin{algorithmic}[1]
\Require The training samples $\mathcal{D}^{s}=\left\{\left(x_{i}^{s}, y_{i}^{s}\right)\right\}_{i=1}^{N^s}$, the pretrained backbone (ViT-base \cite{dosovitskiy2020image}), the maximum iteration $\mathrm{max}_{iter}$, the iteration of the update prototype $\mathrm{update}_{iter}$, trade-off parameter $\eta$, the RMSprop optimizer for the visual-to-semantic mapping network and the Adam optimizer for curriculum learning. The initialized model parameters $\theta^0$ and $\varphi^0$, which denote the training parameters in the visual-to-semantic mapping network and PCL, respectively.
\Ensure 
\While{$iter \leq \mathrm{max}_{iter}$}:
\State Take backbone to extract the visual features $\{v_i(x) \in R^{768 \times 196}\}^{batch}_{i=1}$ in a mini-batch.
\State Visual features are mapped to semantic space with $\theta^0$ in visual-to-semantic mapping network (Eq. \ref{eq:attribute_weight}-Eq. \ref{eq:class_score}) and sample are weighting with $\varphi^0$ in PCL (Eq. \ref{eq:lp}-Eq. \ref{eq:pcl}).
\State Update parameters $\theta^0$ to $\theta^1$ using RMSprop optimizer.
\State In the same step, visual features are mapped to semantic space with $\theta^1$ in visual-to-semantic mapping network (Eq. \ref{eq:attribute_weight}-Eq. \ref{eq:class_score}) and sample are weighting with $\varphi^0$ in PCL (Eq. \ref{eq:lp}-Eq. \ref{eq:pcl}).
\State Update parameters $\varphi^0$ to $\varphi^1$ using Adam optimizer.
\If{$iter \geq \mathrm{update}_{iter}$}:
\State Update the class-level semantic prototypes in PUP (Eq. \ref{eq:prototype_update_seen} and Eq. \ref{eq:prototype_update_unseen}).
\EndIf
\EndWhile
\end{algorithmic}
\end{algorithm}

\subsection{Model Optimization and Zero-Shot Prediction}
In the GZSL setting, the test data includes both seen and unseen classes, and the model tends to overfit to the seen classes during prediction. To address this issue, we introduce a self-calibration loss $\mathcal{L}_{SC}$, which encourages the model to shift some of the predicted probability mass from the seen classes to the unseen classes during training \cite{Huynh2020FineGrainedGZ}:
\begin{gather}
\small
\label{eq:l_sc}\
\begin{aligned}
\mathcal{L}_{SC}=-\frac{1}{n} \sum_{i=1}^{n}   \sum_{c^{\prime=1}}^{\mathcal{C}^u} \log \frac{\exp \left(\psi(x_i^s) \times Z^{c^{\prime}} + \mathbb {I}_{\left[c^{\prime}\in\mathcal{C}^u\right]}\right)}{\sum_{\hat{c} \in \mathcal{C}} \exp \left(\psi(x_i^s) \times Z^{\hat{c}} + \mathbb {I}_{\left[\hat{c}\in\mathcal{C}^u\right]}\right)},
\end{aligned}
\end{gather}
\noindent where $\mathbb {I}_{\left[c\in \mathcal{C}^u\right]}$ is an indicator function that takes the value 1 if $c\in\mathcal{C}^u$ (i.e., the class is unseen), and -1 otherwise. The self-calibration loss $\mathcal{L}_{SC}$ encourages the model to assign a relatively high (non-zero) probability to the true unseen classes when presented with test samples from unseen categories. Finally, the overall optimization objective of CLZSL is formulated as:
\begin{gather}
\label{eq:l_final}
\mathcal{L}_{\text{PCL}} = \mathcal{L}_{\text{CE}} + \eta \mathcal{L}_{SC},
\end{gather}
\noindent where $\eta$ is a hyperparameter that cces the contributions of the cross-entropy loss $\mathcal{L}_{\text{CE}}$ and $\mathcal{L}_{SC}$. To effectively optimize CLZSL, we employ an alternating optimization strategy, as illustrated in Algorithm \ref{algotithm:CLZSL_optimization}.

After training CLZSL, we obtain the semantic embeddings for a test instance $x_i$, denoted as $\psi(x_i)$. To predict its label, an explicit calibration step is then applied, which can be formulated as:
\begin{gather}
\label{eq:prediction}
c^{*}=\arg \max _{c \in \mathcal{C}^u/\mathcal{C}}\psi(x_i) \times Z^{c}+\mathbb {I}_{\left[c\in\mathcal{C}^u\right]},
\end{gather}
\noindent here, $\mathcal{C}^u/\mathcal{C}$ corresponds to the CZSL/GZSL setting respectively.

\begin{table}[ht]
\small
\centering  
\caption{The seen/unseen splits on AWA2, SUN and CUB.}\vspace{-3mm}
\resizebox{1.0\linewidth}{!}{\small
\begin{tabular}{r|c|c|c}
\hline
\multirow{1}{*}{\textbf{Dataset}}&\multirow{1}{*}{\textbf{Train/Test}}&\multirow{1}{*}{\textbf{Seen/unseen}}&\multirow{1}{*}{\textbf{Attribute}}\\
\hline
AWA2 & 23,527/13,795 & 40/10 & 85\\
SUN & 10,320/4,020 & 645/72 & 102\\
CUB & 7,057/4,731 & 150/50 & 312\\
\hline	
\end{tabular} }\vspace{-2mm}
\label{table:dataset} 
\end{table}

\begin{table*}[ht]
\small
\centering  
\caption{Results ~($\%$) of the state-of-the-art CZSL and GZSL modes on AWA2, SUN and CUB, including generative methods, common subspace methods and embedding-based methods. The best and second-best results are marked in \textbf{\color{red}Red} and \textbf{\color{blue}Blue}, respectively. The symbol “--” indicates no results.}\vspace{-3mm}
\resizebox{1.0\linewidth}{!}{\small
\begin{tabular}{r|c|ccc|c|ccc|c|ccc}
\hline
\multirow{3}{*}{\textbf{Methods}} 
&\multicolumn{4}{c|}{\textbf{AWA2}}&\multicolumn{4}{c|}{\textbf{SUN}}&\multicolumn{4}{c}{\textbf{CUB}}\\
\cline{2-5}\cline{6-9}\cline{9-13}
&\multicolumn{1}{c|}{CZSL}&\multicolumn{3}{c|}{GZSL}&\multicolumn{1}{c|}{CZSL}&\multicolumn{3}{c|}{GZSL}&\multicolumn{1}{c|}{CZSL}&\multicolumn{3}{c}{GZSL}\\
\cline{2-5}\cline{6-9}\cline{9-13}
\textbf{} 
&\rm{acc}&\rm{U} & \rm{S} & \rm{H} &\rm{acc}&\rm{U} & \rm{S} & \rm{H} &\rm{acc}&\rm{U} & \rm{S} & \rm{H} \\
\hline
\textbf{Generative Methods} &&&&&&&&&&&&\\ 
\rowcolor{mygray}f-CLSWGAN(CVPR'18)~\cite{Xian2018FeatureGN}&68.2&57.9&61.4&59.6&60.8&42.6&36.6&39.4&57.3&43.7&57.7&49.7\\
f-VAEGAN-D2(CVPR'19)~\cite{Xian2019FVAEGAND2AF}&71.1&57.6&70.6&63.5&64.7&45.1&38.0&41.3&61.0&48.4&60.1&53.6\\
\rowcolor{mygray}E-PGN(CVPR'20)~\cite{Yu2020EpisodeBasedPG}&\textbf{\color{blue}73.4}&52.6&83.5&64.6&--&--&--&--&72.4&52.0&61.1&56.2\\
Composer(NeurIPS'20)~\cite{Huynh2020CompositionalZL}&71.5&62.1&77.3&68.8&62.6&\textbf{\color{blue}55.1}&22.0&31.4&69.4&56.4&63.8&59.9\\
\rowcolor{mygray}GCM-CF(CVPR'21)~\cite{Yue2021CounterfactualZA}&--&60.4&75.1&67.0&--&47.9&37.8&42.2&--&61.0&59.7&60.3\\
FREE(ICCV'21)~\cite{Chen2021FREE}&--&60.4&75.4&67.1&--&47.4&37.2&41.7&--&55.7&59.9&57.7\\
\rowcolor{mygray}CDL+OSCO(TPAMI'23)~\cite{cavazza2023no}&--&48.0&71.0&57.1&--&32.0&\textbf{\color{red}65.0}&42.9&--&29.0&69.0&40.6\\
CLSWGAN+DSP(ICML'23)~\cite{chen2023evolving}&--&60.0&86.0&70.7&--&48.3&43.0&\textbf{\color{blue}45.5}&--&51.4&63.8&56.9\\
\hline
\textbf{Common Subspace Methods} &&&&&&&&&&&&\\ 
\rowcolor{mygray}DeViSE(NeurIPS'13)~\cite{Frome2013DeViSEAD}&54.2&17.1&74.7&27.8&56.5&16.9&27.4&20.9&52.0&23.8&53.0&32.8\\
DCN(NeurIPS'18)~\cite{Liu2018GeneralizedZL}&65.2&25.5&84.2&39.1&61.8&25.5&37.0&30.2&56.2&28.4&60.7&38.7\\
\rowcolor{mygray}CADA-VAE(CVPR'19)~\cite{Schnfeld2019GeneralizedZA}&63.0&55.8&75.0&63.9&61.7&47.2&35.7&40.6&59.8&51.6&53.5&52.4\\
SGAL(NeurIPS'19)~\cite{Yu2019ZeroshotLV}&--&52.5&86.3&65.3&--&35.5&34.4&34.9&--&40.9&55.3&47.0\\
\rowcolor{mygray}HSVA(NeurIPS'21)~\cite{Chen2021HSVA}&--&59.3&76.6&66.8&63.8&48.6&39.0&43.3&62.8&52.7&58.3&55.3\\
\hline 
\textbf{Embedding-based Methods} &&&&&&&&&&&&\\   
\rowcolor{mygray}SP-AEN(CVPR'18)~~\cite{Chen2018ZeroShotVR}&58.5&23.3&90.9&37.1&59.2&24.9&38.6&30.3&55.4&34.7&70.6&46.6\\
SGMA(NeurIPS'19)~\cite{Zhu2019SemanticGuidedML}&68.8&37.6&87.1&52.5&--&--&--&--&71.0&36.7&71.3&48.5\\
\rowcolor{mygray}AREN(CVPR'19)~\cite{Xie2019AttentiveRE}&67.9&15.6&\textbf{\color{blue}92.9}&26.7&60.6&19.0&38.8&25.5&71.8&38.9&\textbf{\color{blue}78.7}&52.1\\
LFGAA(ICCV'19)~\cite{Liu2019AttributeAF}&68.1&27.0&\textbf{\color{red}93.4}&41.9&61.5&18.5&40.0&25.3&67.6&36.2&\textbf{\color{red}80.9}&50.0\\
\rowcolor{mygray}DAZLE(CVPR'20)~\cite{Huynh2020FineGrainedGZ}&67.9&60.3&75.7&67.1&59.4&52.3&24.3&33.2&66.0&56.7&59.6&58.1\\
APN(NeurIPS'20)~\cite{Xu2020AttributePN}&68.4&57.1&72.4&63.9&61.6& 41.9&34.0&37.6&72.0&65.3&69.3&67.2\\
\rowcolor{mygray}SR2E(AAAI'21)~\cite{ge2021semantic}&--&58.0&80.7&67.5&--&43.1&36.8&39.7&--&61.6&70.6&65.8\\
APN+VGSE-SMO(CVPR'22)~\cite{xu2022vgse}&64.0&51.2&81.8&63.0&38.1&24.1&31.8&27.4&28.9&21.9&45.5&29.5\\
\rowcolor{mygray}MSDN(CVPR'22)~\cite{chen2022msdn}&70.1&62.0&74.5&67.7&65.8&52.2&34.2&41.3&76.1&68.7&67.5&68.1\\
Transzero(AAAI'22)~\cite{chen2022transzero}&70.1&61.3&82.3&70.2&65.6&52.6&33.4&40.8&76.8&69.3&68.3&68.8\\
\rowcolor{mygray}DFTN(Neurocomputing'23)~\cite{jia2023dual}&71.2&61.1&78.5&68.7&--&--&--&--&74.2&61.8&67.2&64.4\\
HRT(PR'23)~\cite{cheng2023hybrid}&67.3&58.9&78.7&67.4&63.9&53.2&26.9&35.7&71.7&62.1&63.5&62.8\\
\rowcolor{mygray}CC-ZSL(TCSVT'23)~\cite{cheng2023discriminative}&68.8&62.2&83.1&\textbf{\color{blue}71.1}&62.4&44.4&36.9&40.3&74.3&66.1&73.2&69.5\\
Transzero+ALR(SCIS'24)~\cite{chen2024rethinking}&71.2&61.6&82.3&70.5&66.2&52.7&34.0&41.3&\textbf{\color{red}78.8}&\textbf{\color{red}70.4}&69.0&\textbf{\color{blue}69.7}\\
\rowcolor{mygray}VSGMN(TNNLS'24)~\cite{duan2024visual}&71.2&\textbf{\color{blue}64.0}&77.8&70.3&\textbf{\color{blue}66.3}&50.7&34.1&40.8&77.8&\textbf{\color{blue}69.6}&68.9&69.3\\
DTDR(NeuralNetworks'24)~\cite{zhang2024generalized}&--&\textbf{\color{red}64.1}&76.6&69.8&--&--&--&--&--&54.8&59.5&57.1\\
\rowcolor{mygray}ViFR(IJCV'25)~\cite{chen2025semantics}&\textbf{\color{red}73.7}&58.4&81.4&68.0&65.6&48.8&35.2&40.9&69.1&57.8&62.7&60.1\\
\cdashline{1-13}[4pt/1pt]
{\textbf{CLZSL}}{~\textbf{(Ours)}}&67.4&61.7&85.3&\textbf{\color{red}71.6}&\textbf{\color{red}72.3}&\textbf{\color{red}60.0}&\textbf{\color{blue}49.2}&\textbf{\color{red}54.1}&\textbf{\color{blue}78.0}&67.9&73.1&\textbf{\color{red}70.4}\\
\hline	
\end{tabular} }\vspace{-5mm}
\label{table:final_result} 
\end{table*}

\section{Experiment and evaluation}
\subsection{Datasets and Evaluation Protocols}
\subsubsection{Datasets} We evaluate our method on three challenging benchmark datasets: AWA2 (Animals with Attributes 2) \cite{Xian2017ZeroShotLC}, SUN (SUN Attribute) \cite{Patterson2012SUNAD} and CUB (Caltech UCSD Birds 200) \cite{Welinder2010CaltechUCSDB2}. Among these, SUN and CUB are fine-grained datasets, containing 14,340 and 11,788 images, respectively, while AWA2 is a coarse-grained dataset with 37,322 images. Following the protocol in \cite{Xian2017ZeroShotLC}, we use the same split of seen and unseen classes as shown in Tab. \ref{table:dataset}.

\subsubsection{Evaluation Protocols} Following Xian et al. \cite{Xian2017ZeroShotLC}, we evaluate the top-one accuracy on unseen classes in the CZSL setting, denoted as $acc$. For the GZSL setting, we compute the top-one accuracy on both seen and unseen classes separately. In addition, we also use their harmonic mean to evaluate overall performance in the GZSL setting:
\begin{gather}
\label{eq:harmonic_mean}
H = \frac{2 \times acc_{seen} \times acc_{unseen}}{acc_{seen} + acc_{unseen}}.
\end{gather}

\subsection{Implementation Details}
We divide a $224\times224$ image into $14\times14$ patches of equal size and extract their features $v_i(x) \in R^{768 \times 14 \times 14}$ using the pre-trained ViT-base model \cite{dosovitskiy2020image} on ImageNet. These features $v_i(x)$ are treated as regional features and fed into the visual-to-semantic mapping network. Our model is trained for 50 epochs with a batch size of 50 across all datasets. For optimization, we use the RMSprop optimizer for the visual-to-semantic mapping network and the Adam optimizer for the curriculum learning module. Specifically, for RMSprop, we set the learning rate to 0.0005 for AWA2 and CUB, and 0.0003 for SUN. For Adam, we set the learning rate to 0.01 for all datasets. The trade-off parameter $\eta$ in Eq. \ref{eq:l_final} is set to 0.1, 0.0001, and 0.08 for AWA2, SUN, and CUB, respectively. To enhance the effectiveness of our PUP module, we adopt a strategy whereby prototype update is initiated after the model has been trained for a certain number of epochs. Specifically, we begin updating the manually defined semantic prototypes starting from the 25th epoch for AWA2 and SUN, and the 15th epoch for CUB. All experiments were conducted on a single NVIDIA GeForce RTX 4090 with 24-GB memory, using PyTorch for implementation.

\subsection{Comparison with State of the Arts}
It should be noted that our CLZSL is an embedding-based method. To demonstrate its effectiveness and advantages, we compare its performance with other previous state-of-the-art methods (including embedding-based ZSL methods, generative ZSL methods, common subspace ZSL methods) under both CZSL and GZSL settings.

\subsubsection{Conventional Zero-Shot Learning} We first compare our method with existing state-of-the-art approaches under the CZSL setting. As shown in Tab. \ref{table:final_result}, under the CZSL setting, generative models generally outperform embedding-based models on coarse-grained datasets such as AWA2. This is because generative models can synthesize additional data, enabling better fitting of the underlying data distribution. However, these models perform poorly on fine-grained datasets. The reason is that the generated samples often fail to capture the critical fine-grained features, leading to inferior performance compared to embedding-based models on datasets like SUN and CUB. Our CLZSL achieves 72.3\% and 78.0\% accuracy on the fine-grained datasets SUN and CUB, respectively, and also obtains 67.4\% accuracy on the coarse-grained dataset AWA2. Notably, on the SUN dataset, CLZSL outperforms other methods by at least 6.0\%. These results demonstrate that CLZSL can effectively leverage knowledge learned from seen classes to recognize unseen classes. The advantage of our method lies in the designed curriculum learning strategy, which prioritizes samples with high cosine similarity between their visual mappings and the class-level semantic prototypes, and progressively advances less-aligned samples, reducing the interference of instance-level mismatches to achieve accurate visual-semantic mapping. Additionally, our prototype update module addresses the class-level imprecision through the update of class-level semantic prototypes, thereby further improving the visual-semantic mapping.

\subsubsection{Generalized Zero-Shot Learning} Next, we compare our method with existing state-of-the-art approaches under the GZSL setting, where both seen and unseen classes are considered during prediction. The results are summarized in Tab. \ref{table:final_result}, where $\rm{U}$ denotes the test accuracy on unseen classes, $\rm{S}$ represents the test accuracy on seen classes, and $\rm{H}$ indicates the harmonic mean of the two. As shown in the table, CLZSL achieves improved harmonic mean scores across all three datasets. Specifically, the harmonic mean is enhanced by at least 0.5\%, 8.6\%, and 0.7\% for AWA2, SUN, and CUB, respectively. Moreover, many existing methods achieve substantially higher accuracy on seen classes compared to unseen classes, indicating poor generalization to unseen categories. In contrast, our method not only balances the performance between seen and unseen classes but also enhances the model’s generalization ability on unseen classes. The improvement stems from our curriculum learning strategy, which mitigates instance-level semantic mismatches (e.g., due to occlusion and annotation bias) through progressive learning. Such mismatches may cause the model to rely on irrelevant background features, which can also appear in unseen class samples, leading to bias from seen to unseen classes in GZSL. The PCL module addresses this by progressively enhancing generalization and balancing performance across seen and unseen classes.


\subsection{Ablation Study}
To gain further insights into CLZSL, we conduct ablation studies to evaluate the impact of the curriculum learning module (PCL) and the prototype update module (PUP). In addition, we validate the effectiveness of our approach on different baseline models.

\begin{table}[t]
\small
\centering
\caption{Ablation studies for different components of CLZSL.}\vspace{-3mm}
\setlength{\tabcolsep}{1pt}
\resizebox{0.5\textwidth}{!}
{
\begin{tabular}{l|cccc|cccc|cccc}
\hline
\multirow{2}{*}{\textbf{Methods}} & \multicolumn{4}{c|}{\textbf{AWA2}} & \multicolumn{4}{c|}{\textbf{SUN}} & \multicolumn{4}{c}{\textbf{CUB}} \\
\cline{2-5}\cline{6-9}\cline{10-13}
&$\mathrm{acc}$ &$\mathrm{U}$ &$\mathrm{S}$ &$\mathrm{H}^{\hspace{15pt}}$ &$\mathrm{acc}$ &$\mathrm{U}$ &$\mathrm{S}$ &$\mathrm{H}^{\hspace{15pt}}$ &$\mathrm{acc}$ &$\mathrm{U}$ &$\mathrm{S}$ &$\mathrm{H}^{\hspace{15pt}}$\\
\hline
CLZSL w/o PUP+PCL  &64.9&59.1&82.7&$69.0^{\hspace{13pt}}$&72.6&57.7&47.7&$52.2^{\hspace{13pt}}$&72.9&61.0&72.6&$66.3^{\hspace{13pt}}$\\
CLZSL w/o PUP  &66.6&60.0&86.0&$70.7^{\textcolor{red}{\uparrow1.7}}$&72.4&56.0&50.0&$52.8^{\textcolor{red}{\uparrow0.6}}$&75.8&65.6&72.9&$69.0^{\textcolor{red}{\uparrow2.7}}$\\
CLZSL w/o PCL  &66.4&59.9&85.9&$70.6^{\textcolor{red}{\uparrow1.6}}$&72.3&56.4&50.5&$53.3^{\textcolor{red}{\uparrow1.1}}$&75.2&64.4&74.0&$68.9^{\textcolor{red}{\uparrow2.6}}$\\
CLZSL (full)  &67.4&61.7&85.3&$71.6^{\textcolor{red}{\uparrow2.6}}$&72.3&60.0&49.2&$54.1^{\textcolor{red}{\uparrow1.9}}$&78.0&67.9&73.1&$70.4^{\textcolor{red}{\uparrow4.1}}$\\
\hline
\end{tabular}
}\vspace{-3mm}
\label{table:ablation}
\end{table}

\begin{table}[t]
\small
\centering
\caption{Test the effectiveness of our approach at different baseline. The $\P$ denotes baseline combined with our approach.}\vspace{-3mm}
\setlength{\tabcolsep}{1pt}
\resizebox{0.5\textwidth}{!}
{
\begin{tabular}{l|cccc|cccc|cccc}
\hline
\multirow{2}{*}{\textbf{Methods}}&\multicolumn{4}{c|}{\textbf{AWA2}}&\multicolumn{4}{c|}{\textbf{SUN}}&\multicolumn{4}{c}{\textbf{CUB}}\\
\cline{2-5}\cline{6-9}\cline{10-13}
&$\mathrm{acc}$ &$\mathrm{U}$ &$\mathrm{S}$ &$\mathrm{H}^{\hspace{15pt}}$ &$\mathrm{acc}$ &$\mathrm{U}$ &$\mathrm{S}$ &$\mathrm{H}^{\hspace{15pt}}$ &$\mathrm{acc}$ &$\mathrm{U}$ &$\mathrm{S}$ &$\mathrm{H}^{\hspace{15pt}}$\\
\hline
DAZLE &60.8&57.0&82.5&$67.4^{\hspace{13pt}}$&69.3&59.1&44.8&$51.0^{\hspace{13pt}}$&72.4&60.8&71.6&$65.7^{\hspace{13pt}}$\\
DAZLE$^{\P}$ &64.8&63.9&75.2&$69.1^{\textcolor{red}{\uparrow1.7}}$&69.3&60.2&44.9&$51.4^{\textcolor{red}{\uparrow0.4}}$&73.7&65.8&68.9&$67.3^{\textcolor{red}{\uparrow1.6}}$\\
\hline
MSDN &64.9&59.1&82.7&$69.0^{\hspace{13pt}}$&72.6&57.7&47.7&$52.2^{\hspace{13pt}}$&72.9&61.0&72.6&$66.3^{\hspace{13pt}}$\\
MSDN$^{\P}$ &67.4&61.7&85.3&$71.6^{\textcolor{red}{\uparrow2.6}}$&72.3&60.0&49.2&$54.1^{\textcolor{red}{\uparrow1.9}}$&78.0&67.9&73.1&$70.4^{\textcolor{red}{\uparrow4.1}}$\\
\hline
\end{tabular}
}\vspace{-3mm}
\label{table:different_baseline}
\end{table}




\begin{figure*}[t]
\small
\begin{center}
\includegraphics[scale=0.42]{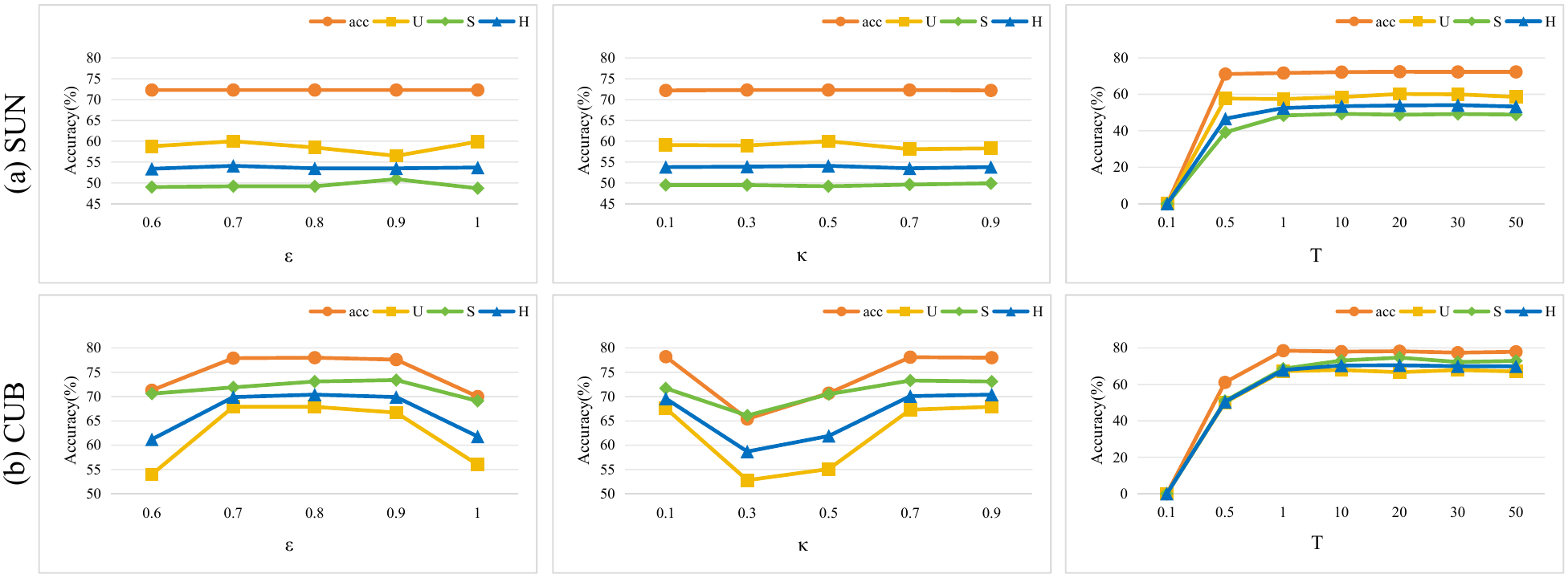}
\\\vspace{-2mm}
\caption{Parameter analysis of PCL on SUN and CUB. The first line is results on SUN, the second line is results on CUB.} \vspace{2mm}
\label{fig:para_PCL}

\includegraphics[scale=0.42]{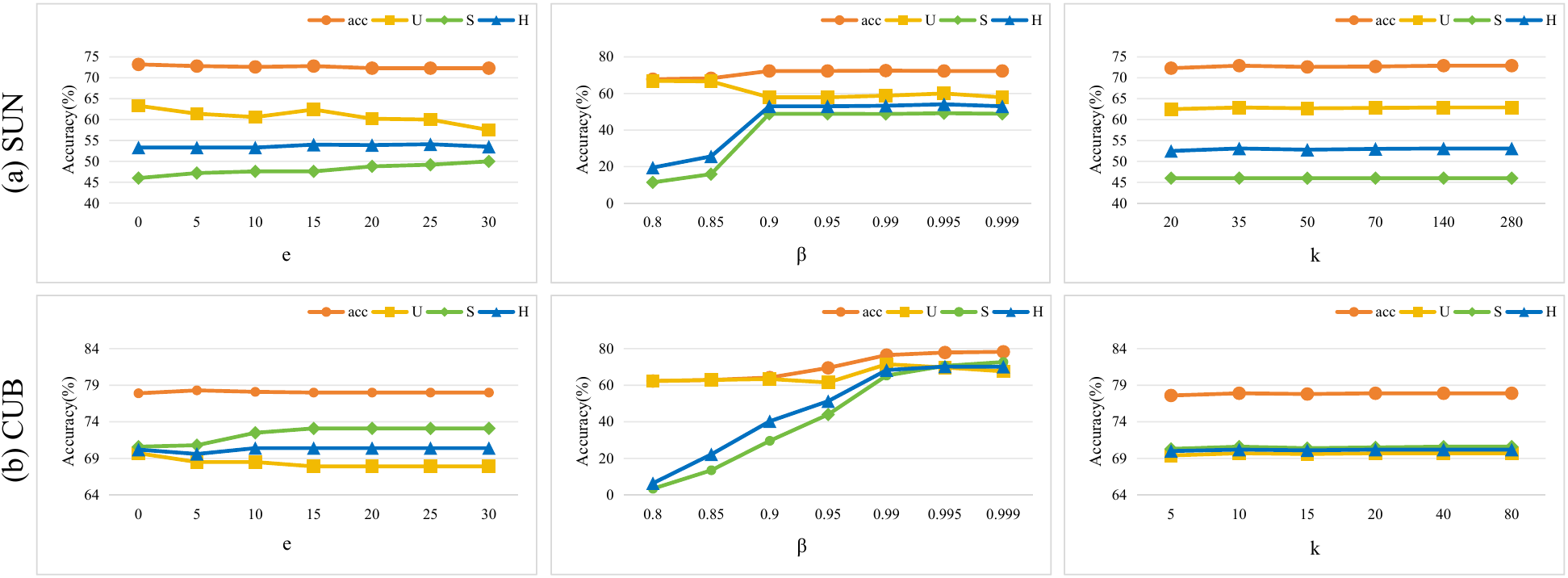}
\\\vspace{-2mm}
\caption{Parameter analysis of PUP on SUN and CUB. The first line is results on SUN, the second line is results on CUB.} \vspace{2mm}
\label{fig:para_PUP}

\includegraphics[scale=0.37]{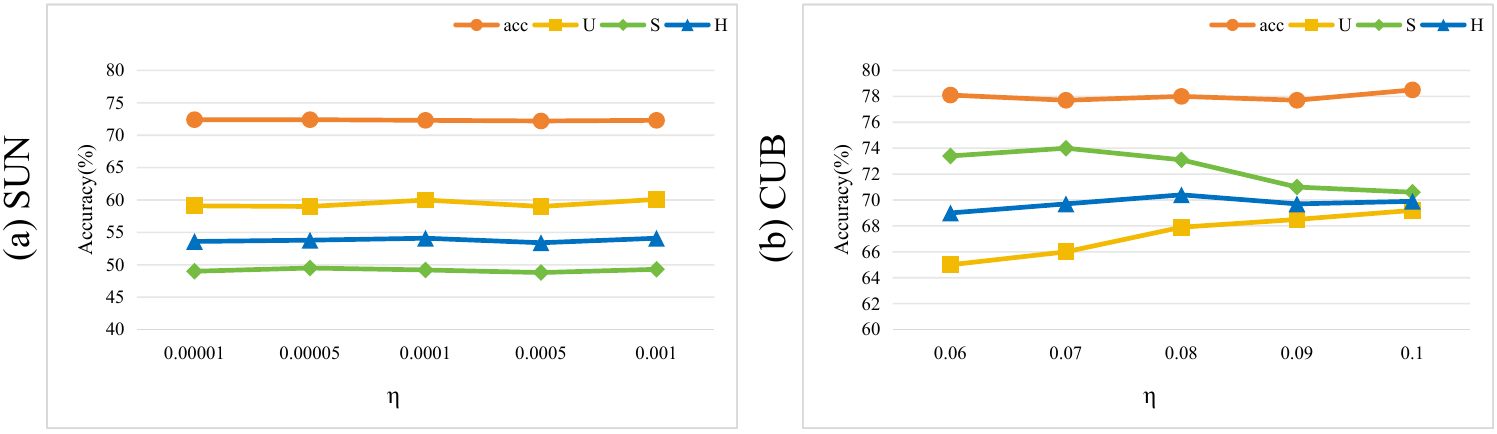}
\\\vspace{-2mm}
\caption{Parameter analysis of $\eta$ on SUN and CUB. The left is results on SUN, the right is results on CUB.}
\vspace{-2mm}
\label{fig:eta}
\end{center}
\end{figure*}

\subsubsection{Analysis of Model Components} We first conduct ablation studies to evaluate the effectiveness of the PCL module (denoted as CLZSL w/o PUP) and the PUP module (denoted as CLZSL w/o PCL). For a fair comparison, all models involved in the ablation experiments are based on features extracted using ViT-base \cite{dosovitskiy2020image}. The results are presented in Tab. \ref{table:ablation}. When the PCL module is used, the harmonic mean scores improve by 1.7\%, 0.6\%, and 2.7\% on AWA2, SUN, and CUB, respectively. This demonstrates the effectiveness of our curriculum learning strategy. Similarly, when the PUP module is employed to update the class-level semantic prototypes, the harmonic means increase by 1.6\%, 1.1\%, and 2.6\% on the same datasets, respectively. This confirms that PUP effectively updates these class-level semantic prototypes and contributes to improved ZSL performance. Furthermore, we observe that combining both the PCL and PUP modules (denoted as CLZSL (full)) leads to even greater improvements. The harmonic mean increases by 2.6\%, 1.9\%, and 4.1\% on AWA2, SUN, and CUB, respectively. This shows that the PUP module can updates class-level semantic prototypes, and enhancing the curriculum learning process with these updated prototypes, thus boosting overall accuracy. Meanwhile, the PCL module enables more precise visual-semantic mapping, which in turn facilitates the update of class-level semantic prototypes. PCL and PUP operate in a mutually reinforcing manner, resulting in enhanced visual-semantic mapping. In addition, since our method is plug-and-play, it has also been evaluated on different baseline models. The results, shown in Tab. \ref{table:different_baseline}, demonstrate that our approach can be integrated with various embedding-based ZSL models and consistently improves their performance.




\begin{figure*}[t]
\small
\begin{center}

\includegraphics[scale=0.52]{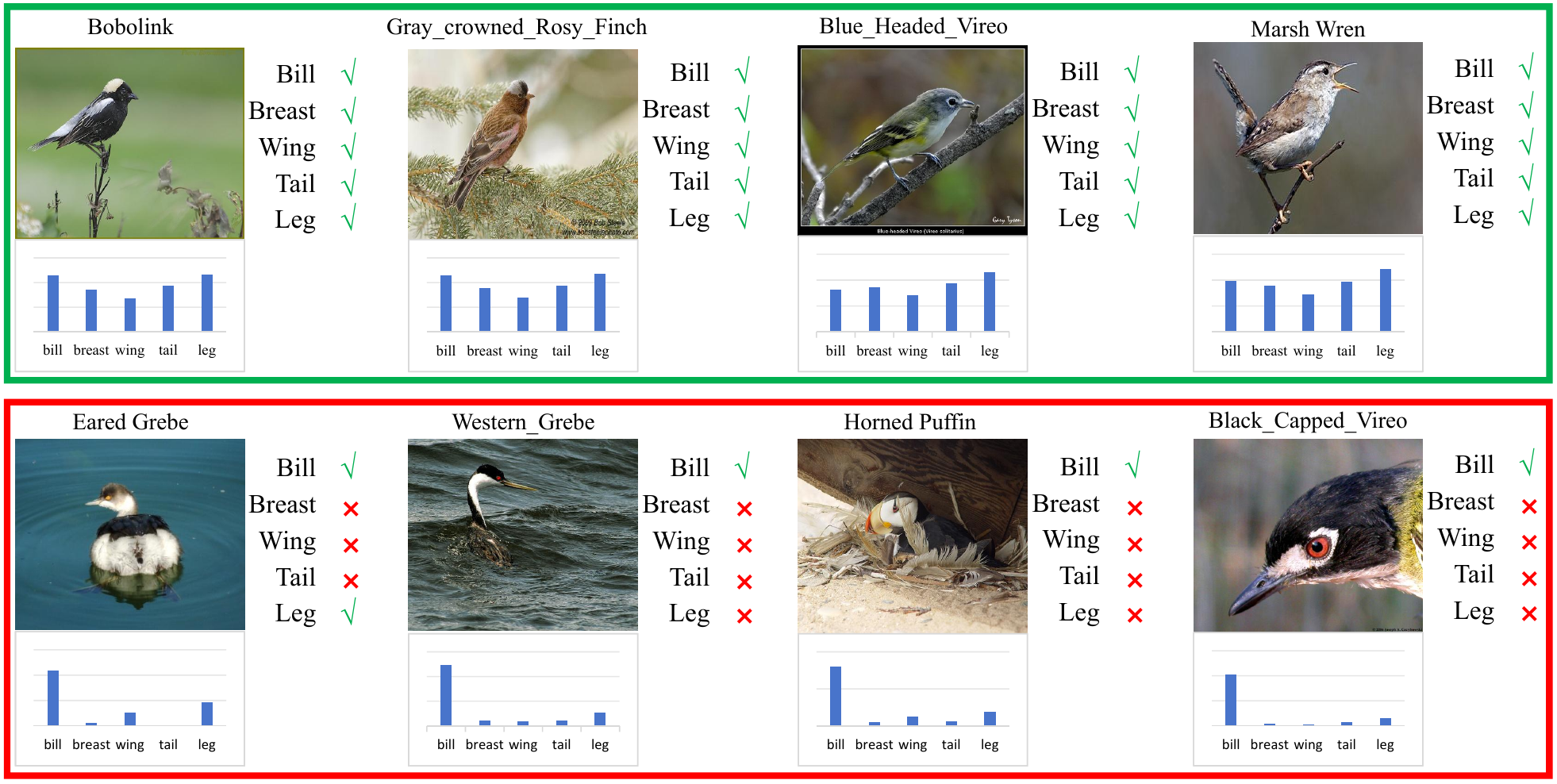}
\\\vspace{-2mm}
\caption{Examples of sample weighting in CLZSL. The green part is the samples with higher weights assigned by the model, and the red part is the samples with lower weights assigned by the model.}
\vspace{2mm}
\label{fig:images_verify}

\includegraphics[scale=0.52]{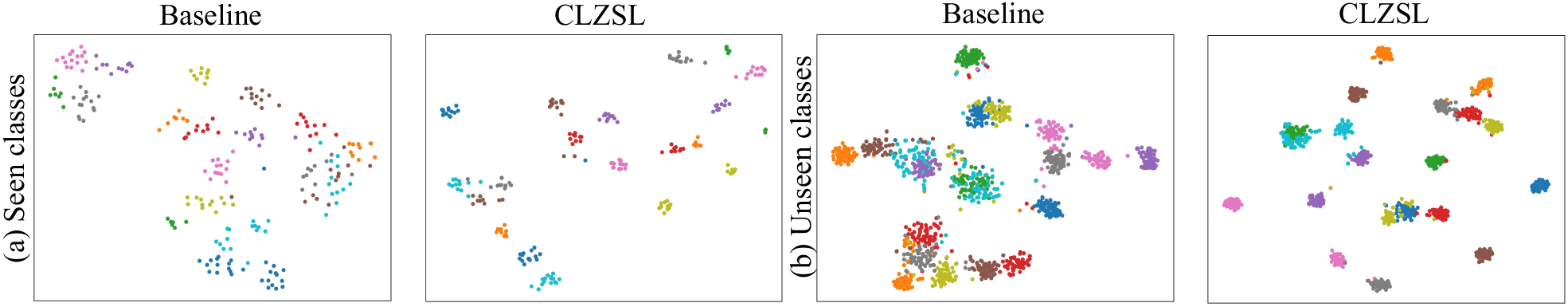}
\\\vspace{-2mm}
\caption{The t-SNE visualization comparison between the baseline model and our method, which describes the embeddings of sample features for 20 seen and unseen classes on CUB, respectively.}
\vspace{-2mm}
\label{fig:tsne}
\end{center}
\end{figure*}

\subsubsection{Hyperparameter analysis of PCL} In the PCL module, the key hyperparameters include the percentile $\varepsilon$ of the exponential moving mean, its decay rate $\kappa$, and the temperature parameter $T$ used in temperature-softmax. The hyperparameter analysis results are shown in Fig. \ref{fig:para_PCL}. First, we evaluate $\varepsilon$ over the range \{0.6, 0.7, 0.8, 0.9, 1.0\}. We find that the highest harmonic mean scores are achieved when $\varepsilon = 0.7$ on SUN and $\varepsilon = 0.8$ on CUB. Next, for the decay rate $\kappa$, we test values in the range \{0.1, 0.3, 0.5, 0.7, 0.9\}. The best performance is obtained with $\kappa = 0.5$ on SUN and $\kappa = 0.9$ on CUB. Additionally, we control the smoothness of the generated sample weights by adjusting the temperature parameter $T$. When $T$ is too small (e.g., $T < 1$), the weight differences between samples become excessively large, causing the model to overfit to a few samples and hindering its ability to learn the overall data distribution. To avoid this, we use a larger $T$, which ensures smoother weight distributions while still allowing the model to distinguish between samples. Based on these findings, we set $T = 30$ for SUN and $T = 10$ for CUB to achieve a balance between effective sample differentiation and stable learning of the data distribution.

\subsubsection{Hyperparameter analysis of PUP} In the PUP module, the main hyperparameters include the training epoch $e$ that determines when the model begins updating the class-level semantic prototypes, the update proportion of the semantic prototypes $\beta$, and the number of selected seen classes $k$ used to update the class-level semantic prototypes of unseen classes. The analysis results are presented in Fig. \ref{fig:para_PUP}. To ensure accurate update of the class-level semantic prototypes, we allow the model to train for a certain number of epochs before initiating the update process. Specifically, we start updating the prototypes at the 25th epoch for SUN and the 15th epoch for CUB. For the number of seen classes $k$, we set it to 35 for SUN and 10 for CUB. This means that we select a number of (approximately 5\% of the total number of categories) seen classes to update the class-level semantic prototypes of unseen classes. Additionally, to prevent over-correction of the class-level semantic prototypes, we set the update proportion $\beta$ to a relatively large value. We evaluate $\beta$ over the range \{0.8, 0.85, 0.9, 0.95, 0.99, 0.995, 0.999\}, and find that the highest harmonic mean scores are achieved when $\beta = 0.995$ for both SUN and CUB.

\subsubsection{Hyperparameter analysis of $\eta$} In CLZSL, we incorporate the self-calibration loss to improve model optimization, where $\eta$ serves as the trade-off parameter balancing the self-calibration loss and the cross-entropy loss. In the following, we analyze the impact of different values of $\eta$. The results are shown in Fig. \ref{fig:eta}. We evaluate $\eta$ over the ranges \{0.00001, 0.00005, 0.0001, 0.0005, 0.001\} for SUN and \{0.06, 0.07, 0.08, 0.09, 0.1\} for CUB. Based on the analysis, we set $\eta = 0.0001$ for SUN and $\eta = 0.08$ for CUB. The results demonstrate that introducing the self-calibration loss helps improve the harmonic mean scores, indicating its effectiveness in enhancing model performance.




\subsection{Qualitative Analysis}
\subsubsection{Examples of sample weighting in CLZSL} To verify the effectiveness of our CLZSL, we select several training samples from the CUB dataset and examine the model's sample weighting behavior by visualizing the weights assigned to these samples. As shown in Fig. \ref{fig:images_verify}, the green-highlighted regions represent samples that are assigned higher weights by the model, while the red-highlighted regions indicate those with lower assigned weights. It can be observed that when the visual features of the target are more distinct (e.g., the green samples), the instance-level semantic prototypes match more accurate with the class-level semantic prototypes, and thus the model assigns them higher weights. In contrast, due to factors such as viewpoint variations, occlusions, and annotation biases, some visual features may be missing or unclear (e.g., the red samples), leading to a less precise match between the instance-level and class-level semantic prototypes. Consequently, the model assigns these samples lower weights. The visualization provides evidence that our model can generate reliable sample weights by leveraging the similarity between visual mappings and class-level semantic prototypes. Combined with previous experimental results, this indicates that our curriculum learning strategy helps the ZSL model effectively address instance-level mismatches, enabling accurate visual-semantic mapping and thereby improving overall model performance.

\subsubsection{The t-SNE Visualization for Visual Embedding} To provide a more intuitive demonstration of the effectiveness of CLZSL, we visualize the t-SNE projections \cite{van2008visualizing} of feature embeddings for samples from the CUB dataset. To enable a clearer comparison, we randomly select 20 categories from both the seen and unseen classes and visualize their corresponding feature embeddings. As shown in Fig. \ref{fig:tsne}, the results compare feature embeddings extracted using the baseline method and CLZSL. It can be observed that CLZSL learns more discriminative feature embeddings for both seen and unseen classes compared to the baseline. This leads to better semantic alignment across different modalities and facilitates more effective knowledge transfer.

\section{DISCUSSION}
Previous ZSL models often ignore the issue that the manually defined semantic prototypes may introduce noisy supervision for two main reasons: (i) instance-level mismatch: variations in perspective, occlusion, and annotation bias will cause discrepancies between individual sample and the
class-level semantic prototypes; and (ii) class-level imprecision: the manually defined semantic prototype may not accurately reflect the true semantics of the class. Consequently, the visual-semantic mapping will be misled, reducing the effectiveness of knowledge transfer to unseen classes. We have proposed a new framework to solve the above problems from two aspects. First, we encourage the model to prioritize samples with high cosine similarity between their visual mappings and the class-level semantic prototypes, and progressively advances less-aligned samples, reducing the interference of instance-level mismatches for better visual-semantic mapping. Second, we dynamically update the class-level semantic prototypes by leveraging the visual mappings learned from instances, improving class-level precision.

Despite its effectiveness, our method still has some limitations: \textbf{1)} The images employed for training the model generally include both semantically relevant and irrelevant features. The inclusion of these semantically irrelevant features can adversely affect the visual-semantic mapping process, thus disrupting the transfer of knowledge from seen to unseen classes. \textbf{2)} Although the current prototype update strategy achieves acceptable performance, it remains relatively simplistic-especially when updating the class-level semantic prototypes for unseen classes. It simply averages visual mappings from seen classes without fully considering their class differences or relationships with unseen classes.


\section{CONCLUSION AND FUTURE WORKS}
In this paper, we propose prototype-guided curriculum learning for Zero-Shot Learning, dubbed as CLZSL. CLZSL mitigates instance-level mismatches through a Prototype-Guided Curriculum Learning (PCL) module and addresses
class-level imprecision via a Prototype Update (PUP) module. Specifically, the PCL module prioritizes samples with high cosine similarity between their visual mappings and the class-level semantic prototypes, and progressively advances less-aligned samples, reducing the interference of instance-level mismatches to achieve accurate visual-semantic mapping. Besides, the PUP module dynamically updates the class-level semantic prototypes by leveraging the visual mappings learned from instances, thereby reducing class-level imprecision and further improving the visual-semantic mapping. To validate the effectiveness of CLZSL, we conduct extensive experiments. The experimental results indicate that our model attains more precise visual-semantic mapping and facilitates more effective knowledge transfer between seen and unseen classes.

Looking ahead, we intend to investigate strategies for mitigating the influence of semantically irrelevant features, potentially leveraging information-theoretic approaches to learn more robust visual representations. Additionally, we plan to explore more sophisticated strategies for prototype updating. For instance, we aim to incorporate the importance of key attributes during the update process, allowing the updated semantic prototypes to better capture inter-class differences and become more discriminative.


\bibliographystyle{IEEEtran} 
\bibliography{ref.bib}

\begin{thebibliography}{10}
\providecommand{\url}[1]{#1}
\csname url@samestyle\endcsname
\providecommand{\newblock}{\relax}
\providecommand{\bibinfo}[2]{#2}
\providecommand{\BIBentrySTDinterwordspacing}{\spaceskip=0pt\relax}
\providecommand{\BIBentryALTinterwordstretchfactor}{4}
\providecommand{\BIBentryALTinterwordspacing}{\spaceskip=\fontdimen2\font plus
\BIBentryALTinterwordstretchfactor\fontdimen3\font minus \fontdimen4\font\relax}
\providecommand{\BIBforeignlanguage}[2]{{%
\expandafter\ifx\csname l@#1\endcsname\relax
\typeout{** WARNING: IEEEtran.bst: No hyphenation pattern has been}%
\typeout{** loaded for the language `#1'. Using the pattern for}%
\typeout{** the default language instead.}%
\else
\language=\csname l@#1\endcsname
\fi
#2}}
\providecommand{\BIBdecl}{\relax}
\BIBdecl

\bibitem{Lampert2014AttributeBasedCF}
C.~H. Lampert, H.~Nickisch, and S.~Harmeling, ``Attribute-based classification for zero-shot visual object categorization,'' \emph{TPAMI}, vol.~36, pp. 453--465, 2014.

\bibitem{Welinder2010CaltechUCSDB2}
P.~Welinder, S.~Branson, T.~Mita, C.~Wah, F.~Schroff, S.~J. Belongie, and P.~Perona, ``Caltech-ucsd birds 200,'' \emph{Technical Report CNS-TR-2010-001, Caltech,}, 2010.

\bibitem{Huang2019GenerativeDA}
H.~Huang, C.~Wang, P.~S. Yu, and C.-D. Wang, ``Generative dual adversarial network for generalized zero-shot learning,'' in \emph{CVPR}, 2019, pp. 801--810.

\bibitem{Li2019LeveragingTI}
J.~Li, M.~Jing, K.~Lu, Z.~Ding, L.~Zhu, and Z.~Huang, ``Leveraging the invariant side of generative zero-shot learning,'' in \emph{CVPR}, 2019, pp. 7394--7403.

\bibitem{pourpanah2022review}
F.~Pourpanah, M.~Abdar, Y.~Luo, X.~Zhou, R.~Wang, C.~P. Lim, X.-Z. Wang, and Q.~J. Wu, ``A review of generalized zero-shot learning methods,'' \emph{TPAMI}, vol.~45, no.~4, pp. 4051--4070, 2022.

\bibitem{jayaraman2014zero}
D.~Jayaraman and K.~Grauman, ``Zero-shot recognition with unreliable attributes,'' \emph{NeurIPS}, vol.~27, 2014.

\bibitem{xu2025information}
H.~Xu, Y.~Gao, J.~Li, and X.~Gao, ``An information compensation framework for zero-shot skeleton-based action recognition,'' \emph{TMM}, 2025.

\bibitem{fu2015zero}
Z.~Fu, T.~Xiang, E.~Kodirov, and S.~Gong, ``Zero-shot object recognition by semantic manifold distance,'' in \emph{CVPR}, 2015, pp. 2635--2644.

\bibitem{zhang2016zero}
Z.~Zhang and V.~Saligrama, ``Zero-shot recognition via structured prediction,'' in \emph{ECCV}, 2016, pp. 533--548.

\bibitem{li2025dcdl}
Q.~Li, S.~Wang, W.~Zhang, S.~Bai, W.~Nie, and A.~Liu, ``Dcdl: Dual causal disentangled learning for zero-shot sketch-based image retrieval,'' \emph{TMM}, 2025.

\bibitem{xu2022attribute}
W.~Xu, Y.~Xian, J.~Wang, B.~Schiele, and Z.~Akata, ``Attribute prototype network for any-shot learning,'' \emph{IJCV}, vol. 130, no.~7, pp. 1735--1753, 2022.

\bibitem{akata2015label}
Z.~Akata, F.~Perronnin, Z.~Harchaoui, and C.~Schmid, ``Label-embedding for image classification,'' \emph{TPAMI}, vol.~38, no.~7, pp. 1425--1438, 2015.

\bibitem{Wang2017ZeroShotVR}
Q.~Wang and K.~Chen, ``Zero-shot visual recognition via bidirectional latent embedding,'' \emph{IJCV}, vol. 124, pp. 356--383, 2017.

\bibitem{Akata2015EvaluationOO}
Z.~Akata, S.~Reed, D.~Walter, H.~Lee, and B.~Schiele, ``Evaluation of output embeddings for fine-grained image classification,'' in \emph{CVPR}, 2015, pp. 2927--2936.

\bibitem{Changpinyo2016SynthesizedCF}
S.~Changpinyo, W.-L. Chao, B.~Gong, and F.~Sha, ``Synthesized classifiers for zero-shot learning,'' in \emph{CVPR}, 2016, pp. 5327--5336.

\bibitem{Yu2020EpisodeBasedPG}
Y.~Yu, Z.~Ji, J.~Han, and Z.~Zhang, ``Episode-based prototype generating network for zero-shot learning,'' in \emph{CVPR}, 2020, pp. 14\,032--14\,041.

\bibitem{Chen2021FREE}
S.~Chen, W.~Wang, B.~Xia, Q.~Peng, X.~You, F.~Zheng, and L.~Shao, ``Free: Feature refinement for generalized zero-shot learning,'' in \emph{ICCV}, 2021.

\bibitem{hou2024visual}
W.~Hou, S.~Chen, S.~Chen, Z.~Hong, Y.~Wang, X.~Feng, S.~Khan, F.~S. Khan, and X.~You, ``Visual-augmented dynamic semantic prototype for generative zero-shot learning,'' in \emph{CVPR}, 2024, pp. 23\,627--23\,637.

\bibitem{Xian2018FeatureGN}
Y.~Xian, T.~Lorenz, B.~Schiele, and Z.~Akata, ``Feature generating networks for zero-shot learning,'' in \emph{CVPR}, 2018, pp. 5542--5551.

\bibitem{Xian2019FVAEGAND2AF}
Y.~Xian, S.~Sharma, B.~Schiele, and Z.~Akata, ``F-vaegan-d2: A feature generating framework for any-shot learning,'' in \emph{CVPR}, 2019, pp. 10\,267--10\,276.

\bibitem{Tsai2017LearningRV}
Y.-H.~H. Tsai, L.-K. Huang, and R.~Salakhutdinov, ``Learning robust visual-semantic embeddings,'' in \emph{ICCV}, 2017, pp. 3591--3600.

\bibitem{Schnfeld2019GeneralizedZA}
E.~Sch{\"o}nfeld, S.~Ebrahimi, S.~Sinha, T.~Darrell, and Z.~Akata, ``Generalized zero- and few-shot learning via aligned variational autoencoders,'' in \emph{CVPR}, 2019, pp. 8239--8247.

\bibitem{He2016DeepRL}
K.~He, X.~Zhang, S.~Ren, and J.~Sun, ``Deep residual learning for image recognition,'' in \emph{CVPR}, 2016, pp. 770--778.

\bibitem{Song2018TransductiveUE}
J.~Song, C.~Shen, Y.~Yang, Y.~Liu, and M.~Song, ``Transductive unbiased embedding for zero-shot learning,'' \emph{CVPR}, pp. 1024--1033, 2018.

\bibitem{Li2018DiscriminativeLO}
Y.~Li, J.~Zhang, J.~Zhang, and K.~Huang, ``Discriminative learning of latent features for zero-shot recognition,'' in \emph{CVPR}, 2018, pp. 7463--7471.

\bibitem{Pennington2014GloveGV}
J.~Pennington, R.~Socher, and C.~D. Manning, ``Glove: Global vectors for word representation,'' in \emph{EMNLP}, 2014.

\bibitem{Xie2019AttentiveRE}
G.-S. Xie, L.~Liu, X.~Jin, F.~Zhu, Z.~Zhang, J.~Qin, Y.~Yao, and L.~Shao, ``Attentive region embedding network for zero-shot learning,'' in \emph{CVPR}, 2019, pp. 9376--9385.

\bibitem{Zhu2019SemanticGuidedML}
Y.~Zhu, J.~Xie, Z.~Tang, X.~Peng, and A.~Elgammal, ``Semantic-guided multi-attention localization for zero-shot learning,'' in \emph{NeurIPS}, 2019.

\bibitem{chen2022msdn}
S.~Chen, Z.~Hong, G.-S. Xie, W.~Yang, Q.~Peng, K.~Wang, J.~Zhao, and X.~You, ``Msdn: Mutually semantic distillation network for zero-shot learning,'' in \emph{CVPR}, 2022, pp. 7612--7621.

\bibitem{chen2022gndan}
S.~Chen, Z.~Hong, G.~Xie, Q.~Peng, X.~You, W.~Ding, and L.~Shao, ``Gndan: Graph navigated dual attention network for zero-shot learning,'' \emph{TNNLS}, vol.~35, no.~4, pp. 4516--4529, 2022.

\bibitem{chen2022transzero}
S.~Chen, Z.~Hong, Y.~Liu, G.-S. Xie, B.~Sun, H.~Li, Q.~Peng, K.~Lu, and X.~You, ``Transzero: Attribute-guided transformer for zero-shot learning,'' in \emph{AAAI}, vol.~36, no.~1, 2022, pp. 330--338.

\bibitem{Xian2017ZeroShotLC}
Y.~Xian, B.~Schiele, and Z.~Akata, ``Zero-shot learning — the good, the bad and the ugly,'' in \emph{CVPR}, 2017, pp. 3077--3086.

\bibitem{Min2020DomainAwareVB}
S.~Min, H.~Yao, H.~Xie, C.~Wang, Z.~Zha, and Y.~Zhang, ``Domain-aware visual bias eliminating for generalized zero-shot learning,'' in \emph{CVPR}, 2020, pp. 12\,661--12\,670.

\bibitem{Han2021ContrastiveEF}
Z.~Han, Z.~Fu, S.~Chen, and J.~Yang, ``Contrastive embedding for generalized zero-shot learning,'' in \emph{CVPR}, 2021.

\bibitem{Chou2021AdaptiveAG}
Y.-Y. Chou, H.-T. Lin, and T.-L. Liu, ``Adaptive and generative zero-shot learning,'' in \emph{ICLR}, 2021.

\bibitem{chen2025semantics}
S.~Chen, Z.~Hong, X.~You, and L.~Shao, ``Semantics-conditioned generative zero-shot learning via feature refinement,'' \emph{IJCV}, pp. 1--18, 2025.

\bibitem{Goodfellow2014GenerativeAN}
I.~J. Goodfellow, J.~Pouget-Abadie, M.~Mirza, B.~Xu, D.~Warde-Farley, S.~Ozair, A.~C. Courville, and Y.~Bengio, ``Generative adversarial nets,'' in \emph{NeurIPS}, 2014.

\bibitem{Mishra2018AGM}
A.~Mishra, M.~K. Reddy, A.~Mittal, and H.~Murthy, ``A generative model for zero shot learning using conditional variational autoencoders,'' in \emph{CVPR Workshop}, 2018, pp. 2269--2277.

\bibitem{Sariyildiz2019GradientMG}
M.~B. Sariyildiz and R.~G. Cinbis, ``Gradient matching generative networks for zero-shot learning,'' in \emph{CVPR}, 2019, pp. 2163--2173.

\bibitem{Wu2020SelfSupervisedDG}
J.~Wu, T.~Zhang, Z.~Zha, J.~Luo, Y.~Zhang, and F.~Wu, ``Self-supervised domain-aware generative network for generalized zero-shot learning,'' in \emph{CVPR}, 2020, pp. 12\,764--12\,773.

\bibitem{Chen2020RethinkingGZ}
Z.~Chen, S.~Wang, J.~Li, and Z.~Huang, ``Rethinking generative zero-shot learning: An ensemble learning perspective for recognising visual patches,'' in \emph{ACM MM}, 2020.

\bibitem{Xie2021GeneralizedZL}
G.-S. Xie, Z.~Zhang, G.-S. Liu, F.~Zhu, L.~Liu, L.~Shao, and X.~Li, ``Generalized zero-shot learning with multiple graph adaptive generative networks.'' \emph{TNNLS}, 2021.

\bibitem{chen2024causal}
S.~Chen, D.~Fu, S.~Chen, S.~Ye, W.~Hou, and X.~You, ``Causal visual-semantic correlation for zero-shot learning,'' in \emph{ACM MM}, 2024, pp. 4246--4255.

\bibitem{hu2025progressive}
L.~Hu, W.~Cao, Z.~Zhang, and Y.~Liang, ``Progressive feature reconstruction network for zero-shot learning,'' \emph{TCSVT}, 2025.

\bibitem{Chen2022DUETCS}
Z.~Chen, Y.~Huang, J.~Chen, Y.~Geng, W.~Zhang, Y.~Fang, J.~Z. Pan, W.~Song, and H.~Chen, ``Duet: Cross-modal semantic grounding for contrastive zero-shot learning,'' in \emph{AAAI}, 2023.

\bibitem{jiang2024dual}
H.~Jiang, Z.~Li, Y.~Hu, B.~Yin, J.~Yang, A.~van~den Hengel, M.-H. Yang, and Y.~Qi, ``Dual prototype contrastive network for generalized zero-shot learning,'' \emph{TCSVT}, 2024.

\bibitem{Xu2020AttributePN}
W.~Xu, Y.~Xian, J.~Wang, B.~Schiele, and Z.~Akata, ``Attribute prototype network for zero-shot learning,'' in \emph{NeurIPS}, 2020.

\bibitem{wang2021dual}
C.~Wang, S.~Min, X.~Chen, X.~Sun, and H.~Li, ``Dual progressive prototype network for generalized zero-shot learning,'' \emph{NeurIPS}, vol.~34, pp. 2936--2948, 2021.

\bibitem{hong2022semantic}
Z.~Hong, S.~Chen, G.-S. Xie, W.~Yang, J.~Zhao, Y.~Shao, Q.~Peng, and X.~You, ``Semantic compression embedding for generative zero-shot learning.'' in \emph{IJCAI}, 2022, pp. 956--963.

\bibitem{chen2022transzero++}
S.~Chen, Z.~Hong, W.~Hou, G.-S. Xie, Y.~Song, J.~Zhao, X.~You, S.~Yan, and L.~Shao, ``Transzero++: Cross attribute-guided transformer for zero-shot learning,'' \emph{TPAMI}, vol.~45, no.~11, pp. 12\,844--12\,861, 2022.

\bibitem{cheng2023discriminative}
D.~Cheng, G.~Wang, N.~Wang, D.~Zhang, Q.~Zhang, and X.~Gao, ``Discriminative and robust attribute alignment for zero-shot learning,'' \emph{TCSVT}, vol.~33, no.~8, pp. 4244--4256, 2023.

\bibitem{vaswani2017attention}
A.~Vaswani, ``Attention is all you need,'' \emph{NeurIPS}, 2017.

\bibitem{jiang2015self}
L.~Jiang, D.~Meng, Q.~Zhao, S.~Shan, and A.~Hauptmann, ``Self-paced curriculum learning,'' in \emph{AAAI}, vol.~29, no.~1, 2015.

\bibitem{wang2021survey}
X.~Wang, Y.~Chen, and W.~Zhu, ``A survey on curriculum learning,'' \emph{TPAMI}, vol.~44, no.~9, pp. 4555--4576, 2021.

\bibitem{han2018co}
B.~Han, Q.~Yao, X.~Yu, G.~Niu, M.~Xu, W.~Hu, I.~Tsang, and M.~Sugiyama, ``Co-teaching: Robust training of deep neural networks with extremely noisy labels,'' \emph{NeurIPS}, vol.~31, 2018.

\bibitem{jiang2018mentornet}
L.~Jiang, Z.~Zhou, T.~Leung, L.-J. Li, and L.~Fei-Fei, ``Mentornet: Learning data-driven curriculum for very deep neural networks on corrupted labels,'' in \emph{ICML}, 2018, pp. 2304--2313.

\bibitem{soviany2022curriculum}
P.~Soviany, R.~T. Ionescu, P.~Rota, and N.~Sebe, ``Curriculum learning: A survey,'' \emph{IJCV}, vol. 130, no.~6, pp. 1526--1565, 2022.

\bibitem{tsvetkov2016learning}
Y.~Tsvetkov, M.~Faruqui, W.~Ling, B.~MacWhinney, and C.~Dyer, ``Learning the curriculum with bayesian optimization for task-specific word representation learning,'' \emph{arXiv preprint arXiv:1605.03852}, 2016.

\bibitem{el2020student}
R.~El-Bouri, D.~Eyre, P.~Watkinson, T.~Zhu, and D.~Clifton, ``Student-teacher curriculum learning via reinforcement learning: Predicting hospital inpatient admission location,'' in \emph{ICML}, 2020, pp. 2848--2857.

\bibitem{fan2017self}
Y.~Fan, R.~He, J.~Liang, and B.~Hu, ``Self-paced learning: An implicit regularization perspective,'' in \emph{AAAI}, vol.~31, no.~1, 2017.

\bibitem{ren2018self}
Z.~Ren, D.~Dong, H.~Li, and C.~Chen, ``Self-paced prioritized curriculum learning with coverage penalty in deep reinforcement learning,'' \emph{TNNLS}, vol.~29, no.~6, pp. 2216--2226, 2018.

\bibitem{platanios2019competence}
E.~A. Platanios, O.~Stretcu, G.~Neubig, B.~Poczos, and T.~M. Mitchell, ``Competence-based curriculum learning for neural machine translation,'' \emph{arXiv preprint arXiv:1903.09848}, 2019.

\bibitem{Huynh2020FineGrainedGZ}
D.~Huynh and E.~Elhamifar, ``Fine-grained generalized zero-shot learning via dense attribute-based attention,'' in \emph{CVPR}, 2020, pp. 4482--4492.

\bibitem{dosovitskiy2020image}
A.~Dosovitskiy, L.~Beyer, A.~Kolesnikov, D.~Weissenborn, X.~Zhai, T.~Unterthiner, M.~Dehghani, M.~Minderer, G.~Heigold, S.~Gelly \emph{et~al.}, ``An image is worth 16x16 words: Transformers for image recognition at scale,'' \emph{arXiv preprint arXiv:2010.11929}, 2020.

\bibitem{Huynh2020CompositionalZL}
D.~T. Huynh and E.~Elhamifar, ``Compositional zero-shot learning via fine-grained dense feature composition,'' in \emph{NeurIPS}, 2020.

\bibitem{Yue2021CounterfactualZA}
Z.~Yue, T.~Wang, H.~Zhang, Q.~Sun, and X.~Hua, ``Counterfactual zero-shot and open-set visual recognition,'' in \emph{CVPR}, 2021.

\bibitem{cavazza2023no}
J.~Cavazza, V.~Murino, and A.~Del~Bue, ``No adversaries to zero-shot learning: Distilling an ensemble of gaussian feature generators,'' \emph{TPAMI}, vol.~45, no.~10, pp. 12\,167--12\,178, 2023.

\bibitem{chen2023evolving}
S.~Chen, W.~Hou, Z.~Hong, X.~Ding, Y.~Song, X.~You, T.~Liu, and K.~Zhang, ``Evolving semantic prototype improves generative zero-shot learning,'' in \emph{ICML}, 2023, pp. 4611--4622.

\bibitem{Frome2013DeViSEAD}
A.~Frome, G.~S. Corrado, J.~Shlens, S.~Bengio, J.~Dean, M.~Ranzato, and T.~Mikolov, ``Devise: A deep visual-semantic embedding model,'' in \emph{NeurIPS}, 2013.

\bibitem{Liu2018GeneralizedZL}
S.~Liu, M.~Long, J.~Wang, and M.~I. Jordan, ``Generalized zero-shot learning with deep calibration network,'' in \emph{NeurIPS}, 2018.

\bibitem{Yu2019ZeroshotLV}
H.~Yu and B.~Lee, ``Zero-shot learning via simultaneous generating and learning,'' in \emph{NeurIPS}, 2019.

\bibitem{Chen2021HSVA}
S.~Chen, G.-S. Xie, Y.~Yang~Liu, Q.~Peng, B.~Sun, H.~Li, X.~You, and L.~Shao, ``Hsva: Hierarchical semantic-visual adaptation for zero-shot learning,'' in \emph{NeurIPS}, 2021.

\bibitem{Chen2018ZeroShotVR}
L.~Chen, H.~Zhang, J.~Xiao, W.~Liu, and S.~Chang, ``Zero-shot visual recognition using semantics-preserving adversarial embedding networks,'' in \emph{CVPR}, 2018, pp. 1043--1052.

\bibitem{Liu2019AttributeAF}
Y.~Liu, J.~Guo, D.~Cai, and X.~He, ``Attribute attention for semantic disambiguation in zero-shot learning,'' in \emph{ICCV}, 2019, pp. 6697--6706.

\bibitem{ge2021semantic}
J.~Ge, H.~Xie, S.~Min, and Y.~Zhang, ``Semantic-guided reinforced region embedding for generalized zero-shot learning,'' in \emph{AAAI}, vol.~35, no.~2, 2021, pp. 1406--1414.

\bibitem{xu2022vgse}
W.~Xu, Y.~Xian, J.~Wang, B.~Schiele, and Z.~Akata, ``Vgse: Visually-grounded semantic embeddings for zero-shot learning,'' in \emph{CVPR}, 2022, pp. 9316--9325.

\bibitem{jia2023dual}
Z.~Jia, Z.~Zhang, C.~Shan, L.~Wang, and T.~Tan, ``Dual-focus transfer network for zero-shot learning,'' \emph{Neurocomputing}, vol. 541, p. 126264, 2023.

\bibitem{cheng2023hybrid}
D.~Cheng, G.~Wang, B.~Wang, Q.~Zhang, J.~Han, and D.~Zhang, ``Hybrid routing transformer for zero-shot learning,'' \emph{Pattern Recognition}, vol. 137, p. 109270, 2023.

\bibitem{chen2024rethinking}
S.~Chen, S.~Chen, G.-S. Xie, X.~Shu, X.~You, and X.~Li, ``Rethinking attribute localization for zero-shot learning,'' \emph{SCIS}, vol.~67, no.~7, p. 172103, 2024.

\bibitem{duan2024visual}
B.~Duan, S.~Chen, Y.~Guo, G.-S. Xie, W.~Ding, and Y.~Wang, ``Visual--semantic graph matching net for zero-shot learning,'' \emph{TNNLS}, 2024.

\bibitem{zhang2024generalized}
C.~Zhang and Z.~Li, ``Generalized zero-shot learning via discriminative and transferable disentangled representations,'' \emph{Neural Networks}, p. 106964, 2024.

\bibitem{Patterson2012SUNAD}
G.~Patterson and J.~Hays, ``Sun attribute database: Discovering, annotating, and recognizing scene attributes,'' in \emph{CVPR}, 2012, pp. 2751--2758.

\bibitem{van2008visualizing}
L.~Van~der Maaten and G.~Hinton, ``Visualizing data using t-sne.'' \emph{JMLR}, vol.~9, no.~11, 2008.

\end{thebibliography}

\vfill

\end{document}